\documentclass[10pt,twocolumn,letterpaper]{article}

\usepackage{arxiv}

\usepackage{float}
\usepackage{bm}
\usepackage{multirow}
\usepackage{makecell}
\usepackage{comment}

\newcommand{\expnumber}[2]{{#1}\mathrm{e}{#2}}

\definecolor{tabfirst}{rgb}{1,0.7,0.7}
\definecolor{tabsecond}{rgb}{1,0.85,0.7}
\definecolor{tabthird}{rgb}{1,1,0.7}
\newcommand{\tabfirst}[1]{\colorbox{tabfirst}{#1}}
\newcommand{\tabsecond}[1]{\colorbox{tabsecond}{#1}}

\definecolor{arxivblue}{rgb}{0.21,0.49,0.74}
\usepackage[breaklinks,colorlinks,allcolors=arxivblue]{hyperref}
\hypersetup{
    pdftitle={ViDS: Video Diffusion Shader using 3D Face Tracking},
    pdfauthor={Wenbo Ji, Davide Davoli, Zhe Chen, Liam Schoneveld, Matthias Niessner, Jiapeng Tang}
}

\newcommand{\diffconditionfigurewidth}{\linewidth}
\newcommand{\referenceshapefigurewidth}{\linewidth}

\title{ViDS: Video Diffusion Shader using 3D Face Tracking}

\author{%
    {\normalsize
    Wenbo Ji$^{1}$ \hspace{0.8em}
    Davide Davoli$^{2}$ \hspace{0.8em}
    Zhe Chen$^{3}$ \hspace{0.8em}
    Liam Schoneveld$^{3}$}\\
    {\normalsize
    Matthias Nie{\ss}ner$^{1}$ \hspace{0.8em}
    Jiapeng Tang$^{1,\dagger}$}\\[3pt]
    {\small
    $^{1}$Technical University of Munich \quad
    $^{2}$Toyota Motor Europe NV/SA \quad
    $^{3}$Woven by Toyota}\\[16pt]
}

\newcommand{\addcorrespondingauthornote}{%
    \AddToShipoutPicture*{%
        \AtPageLowerLeft{%
            \hspace*{\dimexpr 1in+\hoffset+\oddsidemargin\relax}%
            \raisebox{0.60in}[0pt][0pt]{%
                \begin{minipage}{\columnwidth}
                    \footnotesize
                    \rule{0.4\columnwidth}{0.4pt}\\[-2pt]
                    \textsuperscript{$\dagger$}Corresponding author
                \end{minipage}%
            }%
        }%
    }%
}

\makeatletter
\let\arxivstandardparagraph\paragraph
\newcommand{\compactmainpaperlayout}{%
    \renewcommand\paragraph{%
        \@startsection{paragraph}{4}{\z@}%
        {3.0ex \@plus .6ex \@minus .2ex}%
        {-1em}%
        {\normalfont\normalsize\bfseries}%
    }%
    \setlength{\floatsep}{6pt plus 1pt minus 1pt}%
    \setlength{\textfloatsep}{8pt plus 1pt minus 2pt}%
    \setlength{\intextsep}{6pt plus 1pt minus 1pt}%
    \setlength{\dblfloatsep}{6pt plus 1pt minus 1pt}%
    \setlength{\dbltextfloatsep}{10pt plus 1pt minus 2pt}%
    \captionsetup[figure]{skip=4pt}%
    \captionsetup[table]{skip=3pt}%
}
\newcommand{\restorestandardlayout}{%
    \let\paragraph\arxivstandardparagraph
    \setlength{\floatsep}{12pt plus 2pt minus 2pt}%
    \setlength{\textfloatsep}{20pt plus 2pt minus 4pt}%
    \setlength{\intextsep}{12pt plus 2pt minus 2pt}%
    \setlength{\dblfloatsep}{12pt plus 2pt minus 2pt}%
    \setlength{\dbltextfloatsep}{20pt plus 2pt minus 4pt}%
    \captionsetup[figure]{skip=10pt}%
}
\makeatother

\makeatletter
\g@addto@macro\@maketitle{%
    \vspace*{-17mm}
    \begin{center}
        \begin{minipage}{\textwidth}
            \centering
            \includegraphics[width=\linewidth]{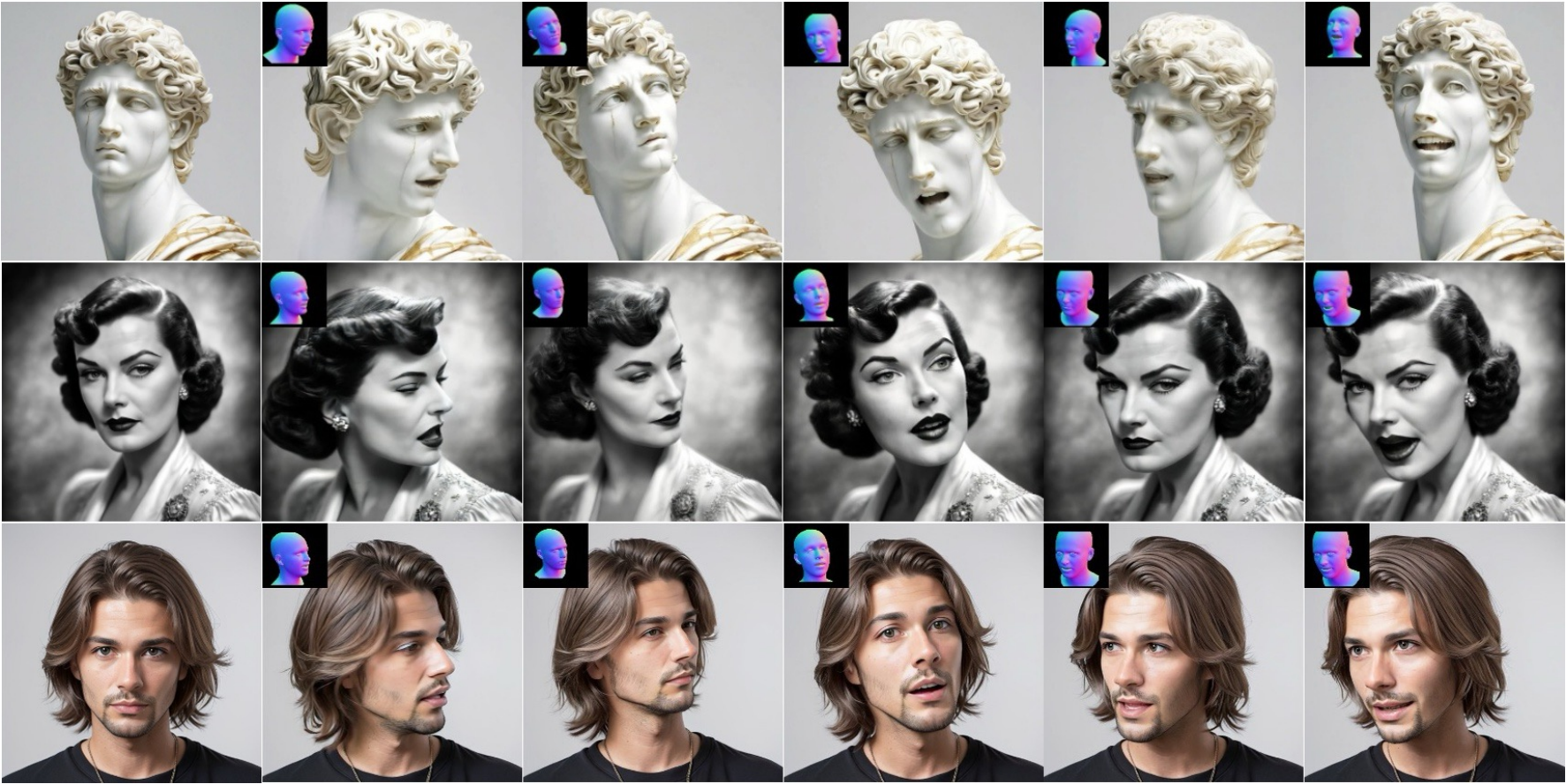}
            \captionsetup{type=figure}
            \captionof{figure}{Given a single face image (first column), our method synthesizes
            vivid facial animations (second to sixth columns) guided by 3DMM tracking
            (top left), including large pose changes and complex expressions.}
            \label{fig:teaser}
        \end{minipage}
    \end{center}
    \vspace*{-1mm}
}
\makeatother

\begin{document}
\addcorrespondingauthornote
\maketitle

\begin{abstract}
We introduce \textbf{ViDS}, a Video Diffusion Shader that leverages 3D face tracking for expressive and identity-preserving portrait animation. We first reconstruct the identity-specific 3DMM mesh from the reference image, and then animate it using expression and pose parameters from a driving video. Leveraging dense geometric cues from 3DMM normal maps, we employ a video diffusion model as a neural shader to synthesize lifelike portrait animations while preserving the appearance and identity of the reference image. We find that more accurate 3DMM tracking enables finer-grained expression control. We also introduce an autoregressive diffusion sampling process that extends generation beyond the model's native window while reducing discontinuities between adjacent clips.
Compared with prior diffusion-based approaches for portrait animation that rely on landmark-based conditioning or implicit motion latents, our method achieves more detailed and consistent expression and pose control while faithfully preserving identity and appearance. Detailed ablation studies validate the effectiveness of our design choices. \href{https://fusheng-ji.github.io/ViDS/}{Project page}.
\end{abstract}


\section{Introduction}
\label{sec:intro}

Portrait animation seeks to synthesize a realistic video of a person under novel poses and expressions from a single reference image. The task requires a difficult balance: the generated video should preserve identity, follow detailed facial motion, and remain temporally stable over long sequences. This balance is especially important for faces, where small geometric or temporal errors are easily perceived.

Classical pipelines fit a 3D Morphable Model (3DMM), transfer pose and expression, and render the animated mesh~\cite{thies2016face2face, thies2015real, thies2018headon, bfm09, li2017learning}. They provide explicit control, but their RGB outputs are limited by the parametric model and simplified shading, especially around the mouth interior, hair, and other non-parametric regions. Recent diffusion-based portrait methods~\cite{ma2024follow-your-emoji, xu2025hunyuanportrait, wei2024aniportrait, chen2025echomimic} improve realism by fine-tuning large pretrained generators~\cite{rombach2022highstable_diffusion, guo2023animatediff, wan2025, xu2025hunyuanportrait}. However, they often rely on sparse landmarks~\cite{lugaresi2019mediapipe} or implicit motion latents~\cite{xu2025hunyuanportrait}, which can be ambiguous under large pose changes, miss subtle expression details, and leak identity cues from the driving video.

We therefore revisit 3DMMs as a geometry-aware conditioning signal for video diffusion. The motivation is practical: 3DMM tracking provides dense, controllable pose and expression information, while video diffusion supplies a strong generative prior for photorealistic appearance. We introduce \textbf{ViDS}, a Video Diffusion Shader that leverages 3D face tracking for expressive and identity-preserving portrait animation. ViDS combines these strengths by reconstructing the reference identity mesh, animating it with driving-video pose and expression parameters, and rendering the result as pixel-aligned normal maps. These normal maps provide local geometric guidance without asking the parametric model to directly render final RGB appearance.
ViDS feeds the reference image, 3DMM normal maps, and noisy video latents into a Wan-based video diffusion model through a simple unified input layer. The reference image anchors identity and appearance, the normal maps drive pose and expression, and a prompt generated from the driving video by a vision-language model (VLM) provides semantic guidance. This design keeps the architecture close to the pretrained video model while making the control signal explicit. For long sequences, each autoregressive window uses the preceding window's generated latent suffix as a fixed prefix; overlapping decoded clips are then blended in pixel space to reduce boundary discontinuities.


We evaluate ViDS on VFHQ~\cite{xie2022vfhq} and Celeb-V-Text. The benchmarks cover self-reenactment and cross-reenactment. The results show that dense 3DMM normal conditioning improves identity preservation and pose and expression control over sparse or implicit alternatives, while the video diffusion prior still synthesizes details outside the 3DMM support, such as inner-mouth appearance and hair motion. Replacing Pixel3DMM with lower-quality tracking substantially degrades identity and geometric fidelity, demonstrating the importance of accurate 3DMM tracking.

\section{Related Work}
\label{sec:related}

\subsection{3D Morphable Face Models}

3D Morphable Face Models (3DMMs) are compact parametric models for facial shape and appearance~\cite{BlanzV99,bfm09}. Later work expanded their training data, expression spaces, and modeled regions~\cite{cao2013facewarehouse, booth20163d, booth20173d, ploumpis2019combining, bolkart2015groupwise, brunton2014multilinear, tran2018nonlinear, li2017learning}, while recent neural variants further improve expressiveness~\cite{giebenhain2023learning, giebenhain2024mononphm, zheng2022imface, yenamandra2021i3dmm, tang2024dphms}.

For monocular images or videos, 3DMM parameters can be obtained by optimization-based fitting~\cite{BlanzV99, wood2022dense, taubner2024flowface, giebenhain2025pixel3dmm, Thies2016Face2FaceRF} or direct regression~\cite{Deng2019Accurate3F, Sanyal2019LearningTR, Feng2020LearningAA, Danek2022EMOCAED, Zielonka2022TowardsMR, Retsinas20243DFE, Schoneveld2025SHeaPSH}. Their low-dimensional parameters still describe dense head pose and expression, making them a useful alternative to sparse landmarks for controllable portrait generation.

\subsection{Neural Rendering}

Neural rendering synthesizes images or videos from explicit or implicit scene representations~\cite{Tewari2020StateOT}. Some methods replace graphics primitives with neural fields or point/mesh representations~\cite{mildenhall2021nerf, park2019deepsdf, kerbl20233d, zhe2022ibrdf, muller2022instant}, while deferred neural rendering keeps geometry but replaces the final shading stage with a network~\cite{zeng2025renderformer, gao2020deferred, thies2019deferred, aliev2020neural}. Generative renderers based on GANs~\cite{Hao2021GANcraftU3} or diffusion models~\cite{DiffusionRenderer} improve visual quality, and video diffusion models further provide temporal priors. Our method follows this neural-rendering view by using 3DMM tracking as geometry and video diffusion as the learned renderer.


\subsection{Talking Head Video Generation}

Talking head generation synthesizes a video of a target identity from motion cues such as audio, text, landmarks, or expression parameters.

\paragraph{Non-diffusion Methods.}
Early methods drive a static face by warping images or feature maps with landmarks or keypoints, as in FOMM~\cite{Siarohin_2019_NeurIPS} and follow-up pseudo-3D variants~\cite{wang2021facevid2vid, Drobyshev22MP, Xu2024VASA1LA}. GAN- or attention-based systems improve identity preservation and lip synchronization~\cite{Zhang2021FlowguidedOT, Zhou2021Pose, zhang2024personatalk}, but the lack of explicit 3D structure makes it difficult to handle large rotations and occluded regions.

\paragraph{Diffusion-based Methods.}
Recent video diffusion models generate temporally coherent frames by denoising latent representations~\cite{Ho2022VideoDM, he2022latent, text2video-zero, xing2023make, chen2023videodreamer}.
Diffusion-based talking head models combine strong generative priors with explicit control signals such as audio embeddings~\cite{SadTalker, abs-2211-12368, GuoCLLBZ21} and landmarks~\cite{ma2024follow-your-emoji, Guo2024LivePortraitEP, ma2024followyourpose, Gan2025HumanDiTPD}.
While these approaches produce realistic renderings with natural motion, they lack explicit control over fine-grained facial expressions and poses.
In contrast, we integrate 3DMMs directly into the diffusion denoising process. This design exposes explicit control over facial expressions, enabling high-fidelity, editable, and geometry-consistent talking head synthesis. Several recent methods~\cite{prinzler2024joker, tang2025gaf} use a 3DMM as a conditioning signal to generate multi-view images. As image-based methods, however, they do not address the temporal consistency required for video generation.

\section{Method}
\label{sec:method}

\paragraph{Backbone and objective.}
We fine-tune Wan~\cite{wan2025}, which combines a causal video VAE, a Diffusion Transformer (DiT)~\cite{DiT}, and flow matching~\cite{flow_matching}. For clean video latent $\mathbf z_0$, noise $\boldsymbol\epsilon$, flow time $\tau\sim\mathcal U(0,1)$, and conditions $\mathbf c$, we train on $\mathbf z_\tau=(1-\tau)\mathbf z_0+\tau\boldsymbol\epsilon$ with
\begin{equation}
\mathcal L=\mathbb E\!\left[\left\|\mathbf v_\theta(\mathbf z_\tau,\mathbf c,\tau)-(\boldsymbol\epsilon-\mathbf z_0)\right\|_2^2\right].
\label{eq:flow_objective}
\end{equation}
At inference, integrating the learned velocity field from $\tau=1$ to $0$ maps noise to a clean video latent.

ViDS synthesizes realistic videos of a source subject from a single image, offering explicit control over facial expressions and head pose via a 3D Morphable Model (3DMM).
Specifically, the generation process is conditioned on a source image $\mathbf{I} \in \mathbb{R}^{H \times W \times 3}$ capturing the subject's identity and appearance, a normal-map sequence $\mathbf{N} \in \mathbb{R}^{T \times H \times W \times 3}$ carrying the desired geometry, and a textual prompt $\mathbf{P}$ generated from the driving video by a vision-language model (VLM).
The resulting video $\mathbf V \in \mathbb{R}^{T\times H \times W \times 3}$ is generated as:
\begin{equation}
    \mathbf{V} = \text{ViDS}\big( \mathbf{I}, \mathbf{N}, \mathbf{P} \big),
\end{equation}
where $T$ denotes the number of frames in the video, and $H$ and $W$ represent the spatial resolution.

\begin{figure*}[t]
    \vspace{-2mm}
    \centering
    \includegraphics[width=\linewidth]{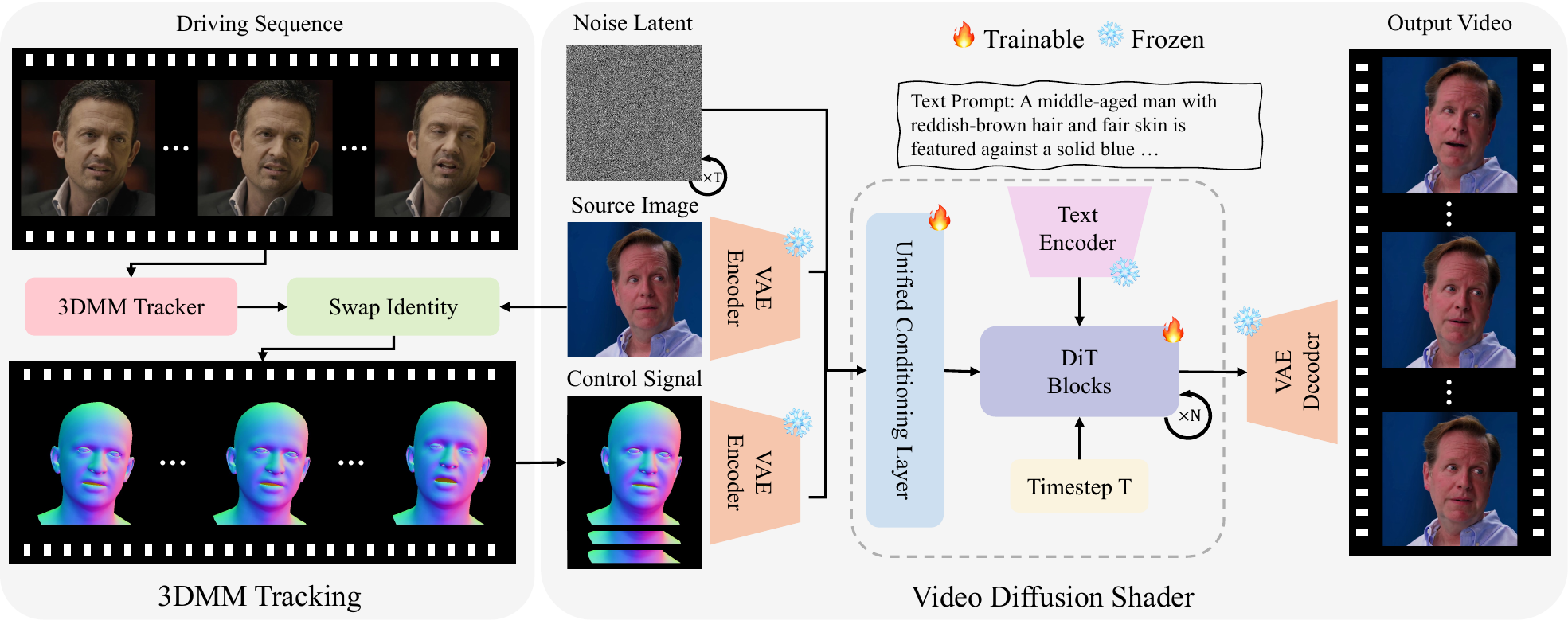}
    \vspace{-4mm}
    \caption{
        \textbf{Pipeline Overview.}
            We introduce a video diffusion shader for controllable portrait animation from a single image. Given a source image and a driving sequence, the goal is to generate a portrait video that preserves the identity and appearance of the source while following the pose and expression changes in the driving video. We achieve this by using 3DMM tracking as the conditioning signal. We first reconstruct a 3DMM mesh from the source image, and then animate this mesh using the pose and expression parameters estimated from the driving sequence. The animated 3DMM is rendered as normal maps, which are concatenated with the source image and the noisy latents before being fed into the DiT model. A VLM-generated prompt derived from the driving video provides an additional condition.
    }
    \label{fig:pipeline}
    \vspace{-1mm}
\end{figure*}

An overview of our pipeline is shown in Fig.~\ref{fig:pipeline}. First, we derive dense, pixel-aligned control signals from 3DMM tracking (Sec.~\ref{subsec:method_3dmm_cond}), rendering normal maps while \emph{freezing} identity to disentangle it from per-frame pose and expression. Next, ViDS (Sec.~\ref{subsec:method_vds}) injects the reference image, condition signal, and text via a unified latent concatenation scheme with multiple classifier-free guidance, unifying identity, expression, and appearance with minimal architectural changes. Finally, Sec.~\ref{subsec:method_diffusion} details inference and long-video generation using overlapping sliding windows for identity and temporal consistency.



\subsection{3DMM Conditioning}
\label{subsec:method_3dmm_cond}
To enable precise control of facial expressions and head poses, the video diffusion process is conditioned on geometric signals derived from a 3D Morphable Model (3DMM). Instead of using low-dimensional 3DMM parameters, we render surface normal maps from the tracked mesh as dense, pixel-aligned, and lighting-agnostic guidance (Fig.~\ref{fig:diff_condition}). In cross-identity reenactment, keeping the reference identity shape fixed while transferring only the driving pose and expression reduces identity leakage from the driving subject.
We adopt Pixel3DMM~\cite{giebenhain2025pixel3dmm} to perform 3D face reconstruction and expression tracking from monocular videos. From the reference image $\mathbf{I}$, we estimate identity parameters $\bm{\beta}$. For each driving frame $k\in\{1,\dots,T\}$, we estimate pose and expression parameters $\bm{\varphi}_k$ together with camera parameters $\bm{\pi}_k$. Pixel3DMM uses predicted 2.5D geometric priors to optimize these parameters for a parametric face model such as FLAME~\cite{FLAME:SiggraphAsia2017}. We keep $\bm{\beta}$ fixed and construct the target mesh $\mathbf{M}_k=\mathbf{M}(\bm{\beta},\bm{\varphi}_k)$, which is rendered under $\bm{\pi}_k$ to obtain $\mathbf{N}_k$. Stacking $\{\mathbf{N}_k\}_{k=1}^T$ produces $\mathbf{N}\in\mathbb{R}^{T\times H\times W\times 3}$. Compared with SHeaP~\cite{Schoneveld2025SHeaPSH}, Pixel3DMM provides more accurate and stable geometric guidance in our evaluation, yielding sharper expression transfer and more consistent head-pose control.

Finally, the normal maps $\mathbf{N}$ together with the reference image $\mathbf{I}$ are fed into ViDS to synthesize the output video. We further investigate alternative 3DMM-based condition signals, including UV maps, gray mesh renderings, and sparse facial landmarks. Pixel-aligned normal maps achieve the best overall performance across the evaluated representations, although individual alternatives perform better on a few metrics. Normal maps capture detailed local facial geometry, enabling more accurate expression and head-pose transfer. The fixed FLAME topology provides consistent surface correspondence across frames, while the rendered normal maps provide dense, image-aligned geometric guidance.

\subsection{Video Diffusion Shader (ViDS)}
\label{subsec:method_vds}
ViDS builds on the video diffusion transformer~\cite{wan2025} and provides explicit control over pose, expression, and semantics by conditioning the generative process.
Given the reference image $\mathbf{I}$ and the normal-map sequence $\mathbf{N}$ described in \cref{subsec:method_3dmm_cond}, we encode both visual inputs with Wan-VAE into latent embeddings $\mathbf{c}_{\mathbf{I}}, \mathbf{c}_{\mathbf{N}}$. In addition, we obtain text embeddings from the VLM-generated prompt $\mathbf{P}$ via a UMT5 text encoder~\cite{chung2023unimax}: $\mathbf{c}_{\text{txt}} = \mathcal{E}_{\text{txt}}(\mathbf{P})$.
The generative process is modeled as a continuous-time flow~\cite{flow_matching} parameterized by a velocity field that is fully conditioned on all the aforementioned control signals:
\begin{equation}
    \label{eq:velocity_field}
    \mathbf{v}_\theta\left(\mathbf z_\tau,
    \mathbf{c}_{\mathbf{I}},
    \mathbf{c}_{\mathbf{N}},
    \mathbf{c}_{\text{txt}},
    \tau\right),
\end{equation}
where $\mathbf{v}_\theta$ is implemented through a DiT~\cite{DiT}.

\paragraph{Unified Conditioning Layer.}
After obtaining the VAE latents, we use a unified conditioning layer instead of cross-attention or ControlNet, as it requires few architectural changes, adds virtually no learnable parameters, and still provides strong geometric control. At each flow time $\tau\in[0,1]$, we concatenate the VAE latents of the reference image ($\mathbf{c}_{\mathbf{I}}$) and the 3DMM normal maps ($\mathbf{c}_{\mathbf{N}}$) with the noisy video latents $\mathbf{z}_\tau$ along the channel dimension, and feed the result into the DiT backbone. In this design, the reference image anchors identity and appearance, the normal maps drive expression and head pose, and the text branch supplies semantic or style guidance through the DiT cross-attention layers.

\paragraph{Multiple Classifier-Free Guidance.}
To enable separate guidance over \emph{identity}, \emph{motion}, and \emph{style}, we adopt Multiple Classifier-Free Guidance~\cite{ho2022classifierfree} in the flow matching framework. During training, each conditioning modality is randomly dropped with probability $10\%$ to enable unconditional predictions. At inference, we compute both the unconditional velocity field and the condition-specific velocity fields, then combine them using guidance scales to control the strength of each modality.

More specifically, at each flow time $\tau$, we evaluate four different versions of the velocity field defined in~\cref{eq:velocity_field}, each with progressively richer conditioning: \emph{unconditional}, \emph{identity-only}, \emph{identity+geometry}, and \emph{identity+geometry+prompt}.
For simplicity, we refer to them as $\mathbf{v}_\tau^u$, $\mathbf{v}_\tau^i$, $\mathbf{v}_\tau^{ig}$, and $\mathbf{v}_\tau^{igp}$, respectively.
These branches are then combined using weighted differences to form the final velocity field:
\begin{equation}
\mathbf{v}_\tau\!=\!\mathbf{v}_\tau^u\!+\!w_i(\mathbf{v}_\tau^{i}\!-\!\mathbf{v}_\tau^u)\!+\!w_g(\mathbf{v}_\tau^{ig}\!-\!\mathbf{v}_\tau^i)\!+\!w_p(\mathbf{v}_\tau^{igp}\!-\!\mathbf{v}_\tau^{ig}),
\end{equation}
where $w_i$, $w_g$, and $w_p$ are the guidance scales for identity, 3DMM geometry, and text, respectively.
Once ODE integration reaches $\tau=0$, we obtain the clean latent representation $\mathbf{z}_0$, which is decoded by the Wan-VAE decoder to produce the final video $\mathbf{V} = \mathcal{D}_{\text{VAE}}(\mathbf{z}_0)$.
We empirically observe that properly balancing the weights of multiple classifier-free guidance terms is important for accommodating different conditioning signals and achieving high-quality face video generation.

\subsection{Autoregressive Diffusion Generation}
\label{subsec:method_diffusion}
\paragraph{Long-video inference.}
We generate long sequences with an overlapping autoregressive sliding-window strategy. We use windows of $s$ frames, a nominal overlap of $f$ frames, and stride $h=s-f$; the final window is tail-aligned. For the first window, we synthesize a local clip conditioned on $(\mathbf{N}_1,\mathbf{P},\mathbf{I})$ and guidance scales $G=(w_i,w_g,w_p)$. For each subsequent window, the generated suffix from the preceding window is VAE-encoded and fixed as the latent prefix during denoising. The decoded windows are merged using linearly weighted blending over their actual overlap.

\paragraph{Identity preservation.}
A single reference frame $\mathbf{I}$ is used for all windows; the same reference embedding is reused to anchor identity while control signals drive pose and expression. We also set $\mathbf{V}[0]=\mathbf{I}$ at the end so that the first output frame exactly matches the reference image.

\paragraph{Temporal consistency.}
Consecutive windows nominally overlap by $f$ frames, but tail alignment may increase the final overlap. We therefore compute each actual overlap from the window indices. After decoding $\hat{\mathbf{V}}_i$, we linearly blend its overlapping region in pixel space with the accumulated output and directly copy its non-overlapping part. The fixed latent prefix promotes causal continuity, while pixel-space overlap-and-add reduces boundary artifacts.

\compactmainpaperlayout
\section{Experiments}
\label{sec:exper}
\renewcommand{\topfraction}{0.92}
\renewcommand{\bottomfraction}{0.85}
\renewcommand{\textfraction}{0.06}
\renewcommand{\floatpagefraction}{0.75}
\setcounter{topnumber}{3}
\setcounter{bottomnumber}{2}
\setcounter{totalnumber}{5}
\paragraph{Datasets.}
We train on monocular videos from NeRSemble~\cite{kirschstein2023nersemble}, HDTF~\cite{zhang2021hdtf}, and VFHQ~\cite{xie2022vfhq}. Following Hallo-3~\cite{xu2024hallo}, long videos are segmented into motion-aware clips and normalized to 25~FPS; after removing clips shorter than 2\,s, we obtain a candidate pool of 10{,}791 HDTF, 13{,}507 NeRSemble, and 15{,}629 VFHQ clips, totaling 39{,}927 clips (maximum durations 10.05\,s, 5.38\,s, and 14.21\,s). The final model is trained on a filtered 30K subset of this pool. We follow the standard protocol and evaluate on the VFHQ test split: self-reenactment uses all 50 test sequences, while cross-reenactment pairs the first frame of one test sequence as the reference identity with a different sequence as the driving video.

\paragraph{Evaluation Protocol.}
In \textbf{self-reenactment}, ground-truth frames are available. Thus we evaluate reconstruction quality using PSNR, SSIM~\cite{wang2004SSIM}, and LPIPS~\cite{zhang2018lpips}. Identity preservation is measured by the cosine similarity (CSIM) between ArcFace feature embeddings~\cite{deng2019arcface} of the generated images and the source frames. To assess expression and pose accuracy, we use the Average Expression Distance (AED) and Average Pose Distance (APD) derived from a 3DMM estimator~\cite{deng2019accurate}, as well as the Average Keypoint Distance (AKD) computed from a facial landmark detector~\cite{bulat2017far}.
In \textbf{cross-reenactment}, ground-truth target frames are not available. Following prior work~\cite{chu2024Gagavatar, xu2025hunyuanportrait, xie2024xportrait, he2025lam}, we report CSIM, AED, and APD to jointly evaluate identity preservation and driving accuracy across different subjects. All metrics are computed at a resolution of $512 \times 512$.

\paragraph{Baselines.}
We compare our method with state-of-the-art talking face generation approaches, including Follow-Your-Emoji~\cite{ma2024follow-your-emoji}, X-Portrait~\cite{xie2024xportrait}, HunyuanPortrait~\cite{xu2025hunyuanportrait}, and Wan-Animate~\cite{cheng2025wananimate}.
Follow-Your-Emoji uses facial landmarks to control head pose and expressions using the U-Net-based video diffusion model AnimateDiff~\cite{guo2023animatediff}. X-Portrait relies on locally warped image patches generated by a feed-forward reenactment method. HunyuanPortrait adopts an implicit representation that encodes facial expression and head pose. Wan-Animate employs spatially aligned skeleton signals and implicit facial features to control body motion and facial expressions.

\paragraph{Implementations.}
Our model is initialized from Wan2.1-T2V-1.3B. Training is conducted on 8 H100 GPUs (80~GB each) with a batch size of 8. We first train for 50k iterations with a learning rate of $\expnumber{8}{-5}$, followed by another 50k iterations with a reduced learning rate of $\expnumber{2}{-5}$.
The total training takes approximately 14 days. During autoregressive video generation, sampling a 49-frame video clip requires about 290\,s on average.

\subsection{Comparison to State-of-the-Art Methods}
\begin{figure*}[!t]
    \centering
    \vspace*{-6mm}
    \includegraphics[width=\textwidth,height=0.84\textheight,keepaspectratio]{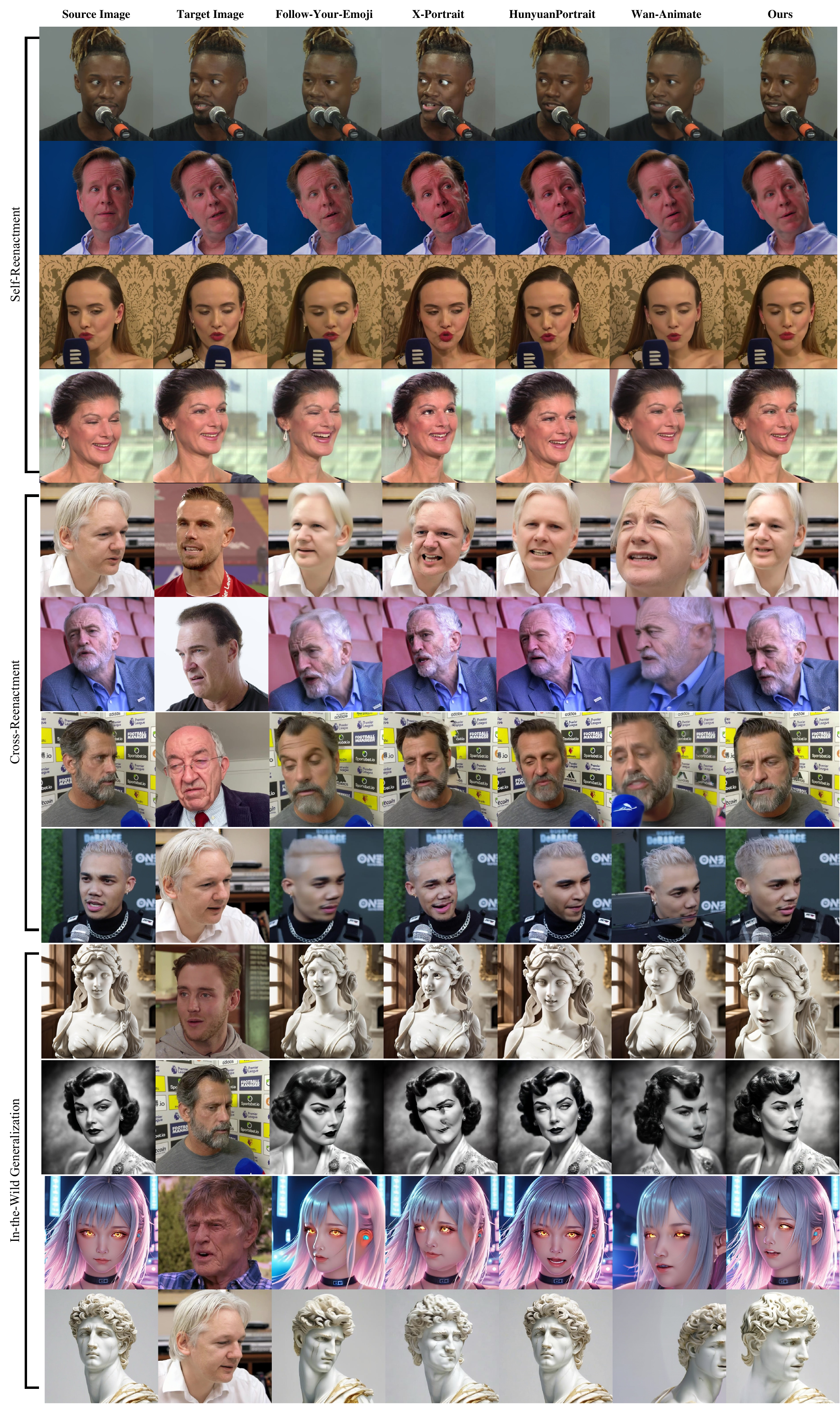}
    {\setlength{\abovecaptionskip}{6pt}
    \caption{
        \textbf{Qualitative comparisons.}
        From top to bottom, we show self-reenactment, cross-reenactment, and in-the-wild generalization. In all three settings, our method outperforms the baselines in handling both extreme head poses and challenging facial expressions, achieving better identity preservation, higher visual quality, and more accurate expression and pose transfer.
    }
    \label{fig:qualitative}
    }
\end{figure*}
\paragraph{Self-Reenactment.}
For each test video, we use the first frame as the reference image and generate the full sequence using the remaining frames as driving images. As shown in the first four rows of Fig.~\ref{fig:qualitative}, our method produces expressions and head poses that are more consistent with the driving video, including eye gaze direction and mouth opening. Our method also outperforms all baselines on most metrics in Tab.~\ref{tab:compare_sota}.


\paragraph{Cross-Reenactment.}
%
As shown in the fifth through eighth rows of Fig.~\ref{fig:qualitative}, the results of X-Portrait~\cite{xie2024xportrait}, Follow-Your-Emoji~\cite{ma2024follow-your-emoji}, and Wan-Animate~\cite{cheng2025wananimate} deviate from the desired expressions. This is because locally warped images and sparse or implicit controls do not provide dense constraints on facial scale and position. In particular, Wan-Animate relies on spatially aligned skeleton signals and implicit facial features; its heuristic bone-length rescaling becomes unreliable under large source--driving shape gaps, causing scale and position misalignment. HunyuanPortrait~\cite{xu2025hunyuanportrait} encodes head pose in an implicit representation but does not expose an explicit, geometrically interpretable pose variable, making cross-identity pose transfer less reliable. It also suffers from identity drift. In contrast, our method uses dense and accurate geometric conditions for both expression and pose, which allows it to generate more plausible face videos while preserving identity.

\begin{table}[H]
    \centering
    \renewcommand\arraystretch{1.15}
    \setlength{\tabcolsep}{2pt}
    \scriptsize
    \resizebox{\columnwidth}{!}{
    \begin{tabular}{lccccccccccccc}
        \toprule
        \multirow{2}{*}{Method} & \multicolumn{9}{c}{Self-Reenactment} & \multicolumn{4}{c}{Cross-Reenactment} \\
        \cmidrule(lr){2-10} \cmidrule(lr){11-14}
         & SSIM$\uparrow$ & LPIPS$\downarrow$ & PSNR$\uparrow$ & CSIM$\uparrow$
         & AED$\downarrow$ & APD$\downarrow$ & AKD$\downarrow$ & FID$\downarrow$ & FVD$\downarrow$
         & CSIM$\uparrow$ & AED$\downarrow$ & APD$\downarrow$ & IQA$\uparrow$ \\
        \midrule
        Follow-Your-Emoji~\cite{ma2024follow-your-emoji}
            & 0.691 & 0.209 & 20.25 & 0.833
            & 0.157 & 0.036 & 4.008 & \tabsecond{22.55} & \tabsecond{0.016}
            & \tabsecond{0.366} & 0.305 & 0.053 & 59.34 \\
        X-Portrait~\cite{xie2024xportrait}
            & 0.683 & 0.227 & 19.46 & 0.853
            & \tabsecond{0.118} & 0.031 & 3.134 & 24.10 & 0.031
            & 0.311 & \tabsecond{0.285} & 0.079 & 58.97 \\
        HunyuanPortrait~\cite{xu2025hunyuanportrait}
            & 0.622 & 0.309 & 17.54 & 0.673
            & \tabfirst{0.113} & 0.043 & 17.47 & 26.20 & 0.018
            & 0.252 & \tabfirst{0.279} & 0.080 & 56.74 \\
        Wan-Animate~\cite{cheng2025wananimate}
            & \tabsecond{0.713} & \tabsecond{0.187} & \tabsecond{21.13} & \tabsecond{0.876}
            & 0.118 & \tabfirst{0.017} & \tabsecond{2.775} & 24.01 & \tabfirst{0.013}
            & 0.227 & 0.286 & \tabsecond{0.053} & \tabfirst{64.51} \\
        \midrule
        Ours
            & \tabfirst{0.722} & \tabfirst{0.178} & \tabfirst{21.19} & \tabfirst{0.879}
            & 0.121 & \tabsecond{0.022} & \tabfirst{2.752} & \tabfirst{21.29} & 0.020
            & \tabfirst{0.372} & 0.298 & \tabfirst{0.049} & \tabsecond{62.85} \\
        \bottomrule
    \end{tabular}
    }
    \caption{
    \textbf{Quantitative comparisons on the VFHQ test set.}
   \tabfirst{Best} and \tabsecond{second-best} are highlighted; rankings use unrounded values.
    }
    \label{tab:compare_sota}
\end{table}

Table~\ref{tab:compare_sota} shows that our method achieves the best self-reenactment SSIM, LPIPS, PSNR, CSIM, AKD, and FID, indicating stronger reconstruction quality and identity preservation. In cross-reenactment, it obtains the best CSIM and APD, reflecting better identity retention and head-pose control across subjects. AED and APD are computed with Deep3DFaceRecon~\cite{deng2019accurate}, as in prior work, but automated reconstruction can misalign with perceived quality; we therefore complement them with a user study in Tab.~\ref{tab:user_study}.
The qualitative comparison also clarifies where the differences arise. Landmark-based control can leave the scale and position of the generated face under-constrained when the source and driving identities have different head shapes. Local warping methods inherit this ambiguity and may distort hairlines or contours under large out-of-plane rotations. Implicit motion latents can recover plausible expressions, but they do not expose an explicit head-pose variable, which makes cross-identity pose transfer less reliable.

\paragraph{In-the-Wild Generalization.}
We further evaluate zero-shot in-the-wild generalization by driving reference images outside our training set (\eg, AI-generated photos, stylized portraits, and virtual characters) with signals from the VFHQ test set, without any in-the-wild fine-tuning. As shown in the last four rows of Fig.~\ref{fig:qualitative}, our method outperforms previous approaches in realism, identity preservation, and expression and pose consistency.

\paragraph{Additional Evaluation on Celeb-V-Text.}
To further validate generalization, we conduct additional comparisons on the Celeb-V-Text dataset against the two strongest baselines. As reported in Tab.~\ref{tab:celebxtext}, our method again achieves the best overall performance, particularly in cross-reenactment identity preservation (CSIM) and pose accuracy (APD), confirming that our advantage is not specific to the VFHQ benchmark.
\begin{table}[!htbp]
    \centering
    \renewcommand\arraystretch{1.15}
    \setlength{\tabcolsep}{2pt}
    \scriptsize
    \resizebox{\columnwidth}{!}{
    \begin{tabular}{lccccccccccccc}
        \toprule
        \multirow{2}{*}{Method} & \multicolumn{9}{c}{Self-Reenactment} & \multicolumn{4}{c}{Cross-Reenactment} \\
        \cmidrule(lr){2-10} \cmidrule(lr){11-14}
         & SSIM$\uparrow$ & LPIPS$\downarrow$ & PSNR$\uparrow$ & CSIM$\uparrow$
         & AED$\downarrow$ & APD$\downarrow$ & AKD$\downarrow$ & FID$\downarrow$ & FVD$\downarrow$
         & CSIM$\uparrow$ & AED$\downarrow$ & APD$\downarrow$ & IQA$\uparrow$ \\
        \midrule
        HunyuanPortrait~\cite{xu2025hunyuanportrait}
            & 0.648 & 0.367 & 17.80 & 0.684
            & \tabfirst{0.161} & 0.040 & 16.67 & 63.58 & 0.056
            & \tabsecond{0.308} & \tabfirst{0.298} & \tabsecond{0.060} & 41.59 \\
        Wan-Animate~\cite{cheng2025wananimate}
            & \tabsecond{0.661} & \tabsecond{0.350} & \tabsecond{19.01} & \tabsecond{0.781}
            & \tabsecond{0.172} & \tabsecond{0.037} & \tabfirst{7.403} & \tabsecond{44.51} & \tabfirst{0.034}
            & 0.254 & 0.321 & 0.062 & \tabfirst{50.49} \\
        Ours
            & \tabfirst{0.668} & \tabfirst{0.346} & \tabfirst{19.98} & \tabfirst{0.793}
            & 0.189 & \tabfirst{0.034} & \tabsecond{7.543} & \tabfirst{43.54} & \tabsecond{0.038}
            & \tabfirst{0.584} & \tabsecond{0.306} & \tabfirst{0.038} & \tabsecond{48.14} \\
        \bottomrule
    \end{tabular}
    }
    \caption{
        \textbf{Quantitative comparisons on the Celeb-V-Text dataset.}
        \tabfirst{Best} and \tabsecond{second-best} are highlighted; rankings use unrounded values.
    }
    \label{tab:celebxtext}
\end{table}

\begin{figure}[t]
    \centering
    \begin{subfigure}[t]{0.593\linewidth}
        \centering
        \includegraphics[width=\linewidth]{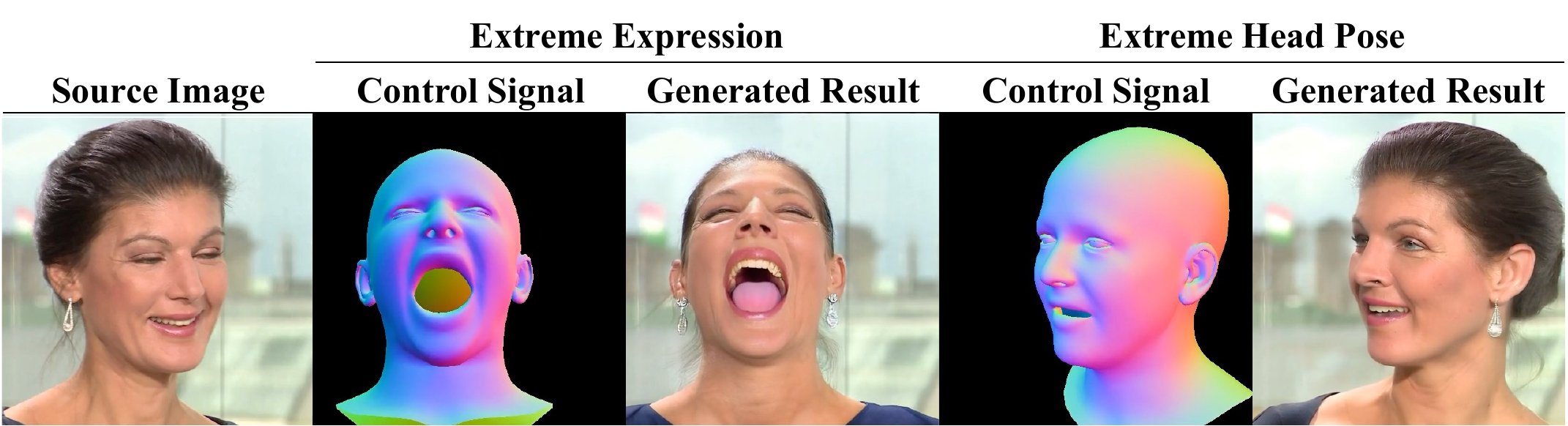}
        \caption{Extreme expression and pose generation.}
        \label{fig:extreme}
    \end{subfigure}
    \hfill
    \begin{subfigure}[t]{0.393\linewidth}
        \centering
        \includegraphics[width=\linewidth]{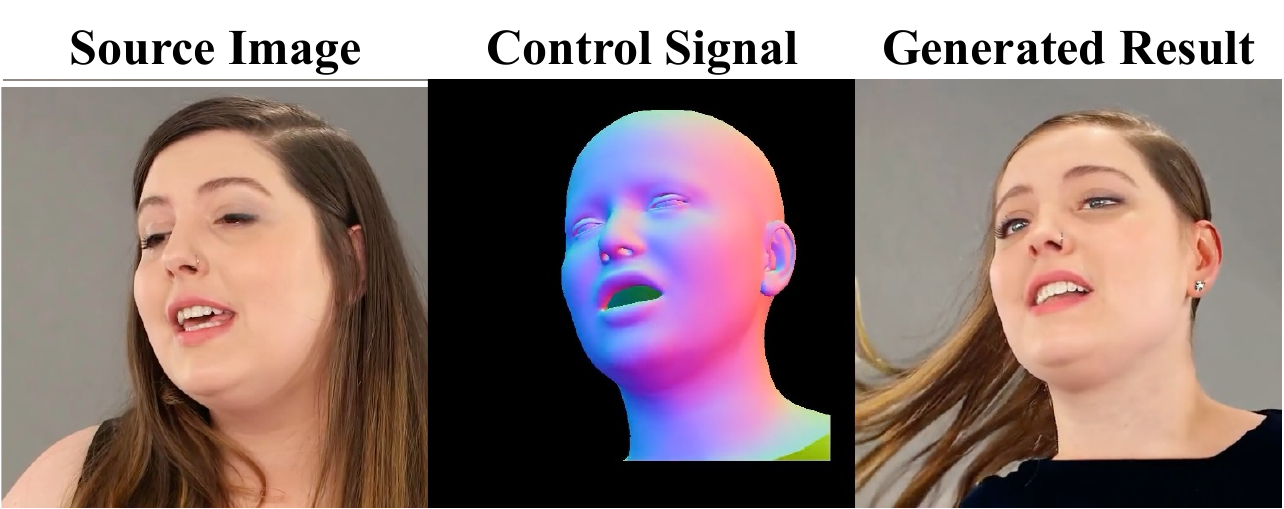}
        \caption{Hair motion under text prompts.}
        \label{fig:hair_motion}
    \end{subfigure}
    \caption{\textbf{Generation beyond rigid 3DMM constraints.} (a) Faithful inner-mouth details, such as teeth and tongue. (b) Text-prompted hair dynamics.}
\end{figure}

\paragraph{Generation Beyond 3DMM Regions.}
Although the 3DMM provides only structural guidance for the face, our diffusion-based shader still synthesizes realistic appearance in non-parametric regions. As shown in Fig.~\ref{fig:extreme} and Fig.~\ref{fig:hair_motion}, our method generates realistic inner-mouth details such as teeth and tongue, handles extreme poses and expressions, and uses text prompts to complement 3DMM guidance for complex hair dynamics.

\paragraph{User Study.}
We conducted a perceptual user study to evaluate the generated videos. We randomly sampled 5 cases from each of the self- and cross-reenactment tasks (10 in total), presented videos of all five methods side-by-side, and collected 400 responses from 40 unique participants. Each video was rated from 1 (worst) to 5 (best) on four aspects: overall quality, temporal smoothness, identity preservation, and expression/pose consistency. As shown in Table~\ref{tab:user_study}, our method achieves the highest scores across all four dimensions in both settings, confirming higher-quality videos with better temporal consistency, more faithful identity preservation, and more accurate motion control.
\begin{table}[!htbp]
    \centering
    \renewcommand\arraystretch{1.3}
    \setlength{\tabcolsep}{2pt}
    \scriptsize
    \resizebox{\columnwidth}{!}{
    \begin{tabular}{lcccccccc}
        \toprule
        \multirow{2}{*}{Method} & \multicolumn{4}{c}{Self-Reenactment} & \multicolumn{4}{c}{Cross-Reenactment} \\
        \cmidrule(lr){2-5} \cmidrule(lr){6-9}
        & \makecell{Overall\\Quality$\uparrow$} & \makecell{Temporal\\Smoothness$\uparrow$} & \makecell{ID\\Preservation$\uparrow$} & \makecell{Expression \&\\Pose Consistency$\uparrow$}
        & \makecell{Overall\\Quality$\uparrow$} & \makecell{Temporal\\Smoothness$\uparrow$} & \makecell{ID\\Preservation$\uparrow$} & \makecell{Expression \&\\Pose Consistency$\uparrow$} \\
        \midrule
        Follow-Your-Emoji~\cite{ma2024follow-your-emoji}
        & 2.62& 2.47& 3.09& 2.75& 2.40& 2.15& 2.73& 2.53\\
        X-Portrait~\cite{xie2024xportrait}
        & 3.25& 3.09& 3.47& 3.23& 2.76& 3.05& 3.04& 2.89 \\
        HunyuanPortrait~\cite{xu2025hunyuanportrait}
        & \tabsecond{3.84}& \tabsecond{3.84}& \tabsecond{3.97}& \tabsecond{3.80}& \tabsecond{3.86}& \tabsecond{3.86}& \tabsecond{3.84}& \tabsecond{3.83}\\
        Wan-Animate~\cite{cheng2025wananimate}
        & 3.58& 3.53& 3.63& 3.59& 3.17& 3.09& 3.12& 3.26\\
        \midrule
        Ours
        & \tabfirst{4.55}& \tabfirst{4.61}& \tabfirst{4.69}& \tabfirst{4.62}& \tabfirst{4.35}& \tabfirst{4.32}& \tabfirst{4.34}& \tabfirst{4.45}\\
        \bottomrule
    \end{tabular}
    }
    \caption{\textbf{Perceptual user study on the VFHQ test set.} Each metric is rated on a scale from 1 (worst) to 5 (best). \tabfirst{Best} and \tabsecond{second-best} results are highlighted; rankings use unrounded values.}
    \label{tab:user_study}
\end{table}

\subsection{Ablation Study}
\paragraph{Fine-tuning strategy.}
As reported in Tab.~\ref{tab:ablation}, full-parameter fine-tuning performs better than LoRA on most reconstruction, identity, and motion metrics, while LoRA obtains better FVD and cross-reenactment IQA\@. Full fine-tuning provides the best overall trade-off, so we use it in the final model.

\paragraph{Conditioning network architecture.}
We adopt a lightweight conditioning layer to inject 3DMM tracking signals into the model. An alternative approach is to use cross-attention layers to fuse 3DMM normal map latents into the DiT backbone. Another option is to introduce additional adaptive layer normalization modules, where the shift and scale parameters are predicted from 3DMM normal latents to modulate intermediate video features.
The comparisons are reported in Tab.~\ref{tab:ablation}. As shown, our conditioning layer achieves better performance than the alternative variants across most metrics in both self- and cross-reenactment settings, while introducing fewer learnable parameters and exhibiting more stable convergence.

\begin{table}[!htbp]
    \centering
    \renewcommand\arraystretch{1.3}
    \setlength{\tabcolsep}{2pt}
    \scriptsize
    \resizebox{\columnwidth}{!}{
    \begin{tabular}{llccccccccccccc}
        \toprule
        \multicolumn{2}{c}{\multirow{2}{*}{Ablation setting}} & \multicolumn{9}{c}{Self-Reenactment} & \multicolumn{4}{c}{Cross-Reenactment} \\
        \cmidrule(lr){3-11} \cmidrule(lr){12-15}
         & & SSIM$\uparrow$ & LPIPS$\downarrow$ & PSNR$\uparrow$ & CSIM$\uparrow$
         & AED$\downarrow$ & APD$\downarrow$ & AKD$\downarrow$ & FID$\downarrow$ & FVD$\downarrow$
         & CSIM$\uparrow$ & AED$\downarrow$ & APD$\downarrow$ & IQA$\uparrow$ \\
        \midrule
        \makecell{Training \\ Strategy}
            & LoRA
            & 0.681 & 0.225 & 19.91 & 0.815
            & 0.165 & 0.032 & 4.152 & 27.91 & \tabsecond{0.016}
            & 0.368 & 0.312 & 0.063 & \tabfirst{64.93} \\
        \midrule
        \multirow{2}{*}{\makecell{Network \\ Structure}}
            & CrossAttn
            & 0.598 & 0.325 & 18.23 & 0.773
            & 0.158 & 0.034 & 3.669 & 46.20 & 0.023
            & 0.344 & 0.292 & 0.098 & 61.52 \\
            & AdaLN
            & 0.674 & 0.228 & 19.59 & 0.829
            & 0.147 & \tabfirst{0.021} & 2.937 & 32.80 & 0.017
            & 0.370 & \tabsecond{0.286} & \tabfirst{0.037} & 58.55 \\
        \midrule
        \multirow{5}{*}{\makecell{Condition \\ Signals}}
            & SHeaP normals
            & 0.642 & 0.286 & 18.70 & 0.750
            & 0.162 & 0.056 & 8.094 & 35.75 & 0.017
            & 0.298 & 0.322 & 0.155 & 57.74 \\
            & Pixel3DMM UV map
            & 0.660 & 0.243 & 19.24 & 0.807
            & 0.149 & 0.035 & 4.055 & 29.83 & 0.036
            & 0.301 & 0.335 & 0.156 & 62.82 \\
            & Pixel3DMM gray mesh
            & 0.656 & 0.254 & 19.38 & 0.816
            & 0.144 & 0.034 & 3.694 & 30.76 & 0.018
            & 0.334 & 0.306 & 0.111 & \tabsecond{63.81} \\
            & Landmark
            & 0.668 & 0.250 & 19.29 & 0.783
            & 0.163 & 0.050 & 5.878 & 28.28 & 0.037
            & 0.334 & 0.310 & 0.080 & 61.45 \\
            & w/o Text
            & \tabsecond{0.714} & \tabsecond{0.188} & \tabsecond{20.88} & \tabsecond{0.862}
            & \tabsecond{0.130} & 0.022 & \tabsecond{2.886} & \tabfirst{21.18} & \tabfirst{0.016}
            & \tabfirst{0.377} & \tabfirst{0.286} & \tabsecond{0.043} & 61.03 \\
        \midrule
        & \textbf{Ours}
            & \tabfirst{0.722} & \tabfirst{0.178} & \tabfirst{21.19} & \tabfirst{0.879}
            & \tabfirst{0.121} & \tabsecond{0.021} & \tabfirst{2.752} & \tabsecond{21.29} & 0.020
            & \tabsecond{0.372} & 0.298 & 0.049 & 62.85 \\
        \bottomrule
    \end{tabular}
    }
    \caption{
    \textbf{Quantitative ablation study on the training strategy, conditioning network architecture, and conditioning signals,} evaluated on the VFHQ test set.
 \tabfirst{Best} and \tabsecond{second-best} are highlighted; rankings use unrounded values.
    }
    \label{tab:ablation}
\end{table}

%

\paragraph{Effect of each conditioning signal.}
We further disable each conditioning signal individually to assess its contribution. As reported in Tab.~\ref{tab:ablation_cfg_alone}, removing the reference image or the 3DMM control signal drastically degrades identity preservation and expression/pose accuracy. Removing text slightly improves cross-reenactment CSIM and FID but reduces the overall semantic and perceptual balance, including IQA\@. The reference and geometry conditions are essential, while the text branch acts as a complementary semantic enhancer.
\begin{table}[H]
    \centering
    \setlength{\tabcolsep}{2pt}
    \scriptsize
    \resizebox{\columnwidth}{!}{
    \begin{tabular}{
        c
        ccccccc
        cc}
        \toprule
        \multirow{2}{*}{Method}
        & \multicolumn{7}{c}{Self-Reenactment}
        & \multicolumn{2}{c}{Cross-Reenactment} \\
        \cmidrule(lr){2-8}
        \cmidrule(lr){9-10}
        & SSIM$\uparrow$ & LPIPS$\downarrow$ & PSNR$\uparrow$ & CSIM$\uparrow$
        & AKD$\downarrow$ & FID$\downarrow$ & FVD$\downarrow$
        & CSIM$\uparrow$ & IQA$\uparrow$ \\
        \midrule
        w/o Image
            & 0.475 & 0.619 & 13.16 & 0.431
            & 25.68 & 215.7 & 0.069
            & 0.294 & 54.07 \\
        w/o Control
            & 0.619 & 0.333 & 17.06 & 0.580
            & 23.83 & 36.82 & 0.150
            & 0.236 & 60.42 \\
        w/o Text
            & \tabsecond{0.719} & \tabsecond{0.179} & \tabsecond{21.08} & \tabsecond{0.875}
            & \tabsecond{2.887} & \tabfirst{21.10} & \tabsecond{0.021}
            & \tabfirst{0.373} & \tabsecond{62.80} \\
        Ours
            & \tabfirst{0.722} & \tabfirst{0.178} & \tabfirst{21.19} & \tabfirst{0.879}
            & \tabfirst{2.752} & \tabsecond{21.29} & \tabfirst{0.020}
            & \tabsecond{0.372} & \tabfirst{62.85} \\
        \bottomrule
    \end{tabular}
    }
    \caption{
        \textbf{Ablation of each conditioning signal on VFHQ.}
        \tabfirst{Best} and \tabsecond{second-best} are highlighted; rankings use unrounded values.
    }
    \label{tab:ablation_cfg_alone}
\end{table}

\begin{figure}[t]
    \centering
    \providecommand{\diffconditionfigurewidth}{0.52\linewidth}
    \includegraphics[width=\diffconditionfigurewidth]{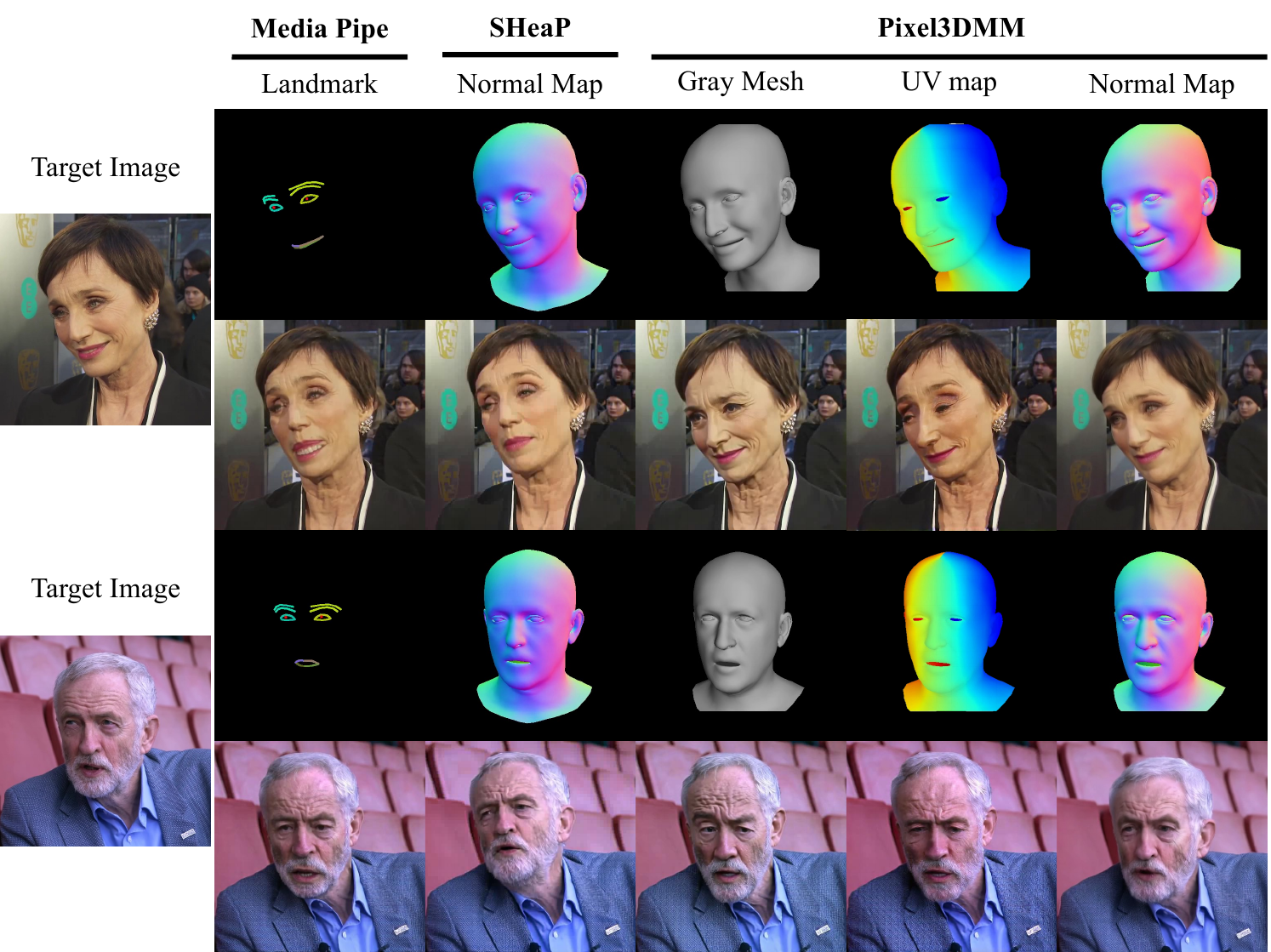}
    \caption{\textbf{Ablation on conditioning signals.} Comparison of different control signals for expression and pose control.}
    \label{fig:diff_condition}
\end{figure}

\paragraph{Types of conditioning signals.}
We evaluate five expression and pose controls: Pixel3DMM normal maps, UV maps, and gray mesh renderings~\cite{giebenhain2025pixel3dmm}; SHeaP normals~\cite{Schoneveld2025SHeaPSH}; and MediaPipe landmarks~\cite{lugaresi2019mediapipe}.
As shown in Fig.~\ref{fig:diff_condition}, UV maps and gray mesh renderings lack fine local surface details, SHeaP normals are less stable under challenging expressions, and landmarks are too sparse for precise control. Pixel3DMM normal maps preserve dense local geometry over a fixed-topology surface, which explains their best overall identity preservation and expression-transfer performance in Tab.~\ref{tab:ablation}, despite a few metric-specific advantages from other representations.

\paragraph{Reference shape parameters.}
The 3DMM shape parameters of the reference image provide critical identity cues. As shown in Fig.~\ref{fig:shape_ablation}, replacing them with driving-video parameters leads to inaccurate proportions, inconsistent contours, and identity mismatch.
\begin{center}
    \providecommand{\referenceshapefigurewidth}{0.92\linewidth}
    \includegraphics[width=\referenceshapefigurewidth]{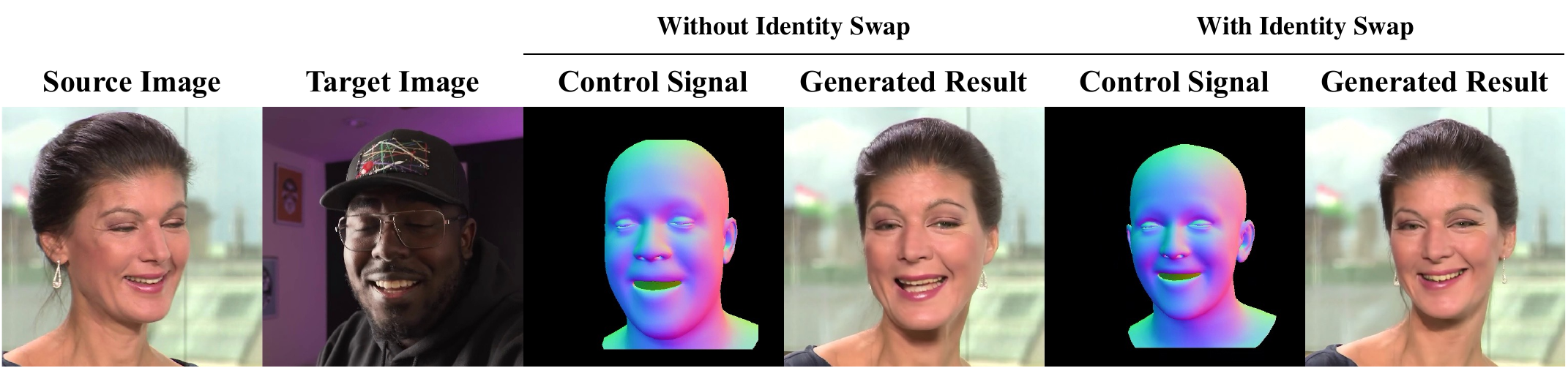}
    \captionof{figure}{\textbf{Ablation on reference shape parameters.} Without identity swapping, using the driving subject's shape parameters produces inaccurate proportions and contours (left). Replacing them with the reference identity parameters restores the correct facial structure (right).}
    \label{fig:shape_ablation}
\end{center}

\paragraph{Classifier-free guidance weights.}
We ablate guidance scales $(w_i,w_g,w_p)$ for identity, geometry, and text. In Tab.~\ref{tab:ablation_cfg}, lowering $w_p$ to 4.0 weakens semantic guidance, whereas raising $w_g$ to 3.5 over-constrains motion; $(2.0,\,1.5,\,6.0)$ gives the best overall trade-off.

\begin{table}[!htbp]
    \centering
    \setlength{\tabcolsep}{2pt}
    \scriptsize
    \resizebox{\columnwidth}{!}{
    \begin{tabular}{
        c c c
        ccccccccc
        cccc}
        \toprule
        \multicolumn{3}{c}{Guidance Scale} & \multicolumn{9}{c}{Self-Reenactment} & \multicolumn{4}{c}{Cross-Reenactment} \\
        \cmidrule(lr){1-3}
        \cmidrule(lr){4-12}
        \cmidrule(lr){13-16}
        \makecell{Ref} &
        \makecell{Ctrl} &
        \makecell{Text} &
        SSIM$\uparrow$ & LPIPS$\downarrow$ & PSNR$\uparrow$ & CSIM$\uparrow$ &
        AED$\downarrow$ & APD$\downarrow$ & AKD$\downarrow$ & FID$\downarrow$ & FVD$\downarrow$ &
        CSIM$\uparrow$ & AED$\downarrow$ & APD$\downarrow$ & IQA$\uparrow$ \\
        \midrule
        2.0 & 1.5 & 4.0
            & 0.659 & 0.240 & 19.39 & \tabsecond{0.817}
            & \tabsecond{0.145} & 0.033 & 3.697 & \tabsecond{27.56} & 0.028
            & 0.338 & 0.305 & 0.102 & \tabsecond{65.28} \\
        2.0 & 3.5 & 6.0
            & \tabsecond{0.663} & \tabsecond{0.238} & \tabsecond{19.49} & 0.816
            & \tabsecond{0.145} & \tabsecond{0.025} & \tabsecond{3.097} & 27.93 & \tabsecond{0.025}
            & \tabsecond{0.361} & \tabfirst{0.280} & 0.067 & \tabfirst{67.06} \\
        2.0 & 1.5 & 6.0
            & \tabfirst{0.722} & \tabfirst{0.178} & \tabfirst{21.19} & \tabfirst{0.879}
            & \tabfirst{0.121} & \tabfirst{0.022} & \tabfirst{2.752} & \tabfirst{21.29} & \tabfirst{0.020}
            & \tabfirst{0.372} & \tabsecond{0.298} & \tabfirst{0.049} & 62.85 \\
        \bottomrule
    \end{tabular}
    }
    \caption{
    \textbf{Ablation of classifier-free guidance scales evaluated on VFHQ.}
 \tabfirst{Best} and \tabsecond{second-best} are highlighted; rankings use unrounded values.
    }
    \label{tab:ablation_cfg}
\end{table}

\paragraph{Training dataset size.}
Tab.~\ref{tab:ablation_size} compares 5K--20K subsets with our final model trained on the filtered 30K subset. Overall performance improves with scale despite metric-level fluctuations, motivating the 30K setting.
\begin{table}[!htbp]
    \centering
    \renewcommand\arraystretch{1.15}
    \setlength{\tabcolsep}{2pt}
    \scriptsize
    \resizebox{\columnwidth}{!}{
    \begin{tabular}{llccccccccccccc}
        \toprule
        \multicolumn{2}{c}{\multirow{2}{*}{Ablation setting}} & \multicolumn{9}{c}{Self-Reenactment} & \multicolumn{4}{c}{Cross-Reenactment} \\
        \cmidrule(lr){3-11} \cmidrule(lr){12-15}
         & & SSIM$\uparrow$ & LPIPS$\downarrow$ & PSNR$\uparrow$ & CSIM$\uparrow$
         & AED$\downarrow$ & APD$\downarrow$ & AKD$\downarrow$ & FID$\downarrow$ & FVD$\downarrow$
         & CSIM$\uparrow$ & AED$\downarrow$ & APD$\downarrow$ & IQA$\uparrow$ \\
        \midrule
        \multirow{4}{*}{\makecell{Dataset \\Size}}
            & \textit{5K}   & 0.711 & 0.183 & 21.02 & 0.872 & 0.129 & 0.019 & \tabfirst{2.669} & \tabsecond{22.23} & \tabfirst{0.009} & 0.370 & \tabsecond{0.282} & \tabfirst{0.036} & 63.23 \\
            & \textit{10K}  & \tabsecond{0.714} & 0.187 & 20.66 & 0.874 & \tabsecond{0.121} & \tabfirst{0.019} & \tabsecond{2.704} & 22.84 & \tabsecond{0.009} & \tabsecond{0.372} & \tabfirst{0.280} & 0.038 & 61.52 \\
            & \textit{15K}  & 0.712 & \tabsecond{0.181} & \tabsecond{21.02} & \tabsecond{0.876} & 0.238 & \tabsecond{0.019} & 2.717 & 22.53 & 0.013 & 0.370 & 0.283 & \tabsecond{0.037} & \tabsecond{63.95} \\
            & \textit{20K} & 0.702 & 0.186 & 20.75 & 0.861 & 0.129 & \tabsecond{0.019} & 2.726 & 23.89 & 0.011 & 0.357 & 0.283 & 0.038 & \tabfirst{64.67} \\
        \midrule
        & \textbf{Ours (30K)}
            & \tabfirst{0.722} & \tabfirst{0.178} & \tabfirst{21.19} & \tabfirst{0.879}
            & \tabfirst{0.121} & 0.021 & 2.752 & \tabfirst{21.29} & 0.020
            & \tabfirst{0.372} & 0.298 & 0.049 & 62.85 \\
        \bottomrule
    \end{tabular}
    }
    \caption{\textbf{Training-size ablation on VFHQ.} \tabfirst{Best} and \tabsecond{second-best} are highlighted based on unrounded values.}
    \label{tab:ablation_size}
\end{table}

\paragraph{Sensitivity to tracking quality.}
Using SHeaP tracking at inference without retraining substantially worsens cross-reenactment CSIM and AKD, while FID and FVD remain comparatively stable (Tab.~\ref{tab:sheap_test}). Accurate tracking is therefore important for identity and geometry control.
\begin{table}[!htbp]
    \centering
    \setlength{\tabcolsep}{2pt}
    \scriptsize
    \resizebox{\columnwidth}{!}{
    \begin{tabular}{
        c
        ccccccc
        cc}
        \toprule
        \multirow{2}{*}{Method}
        & \multicolumn{7}{c}{Self-Reenactment}
        & \multicolumn{2}{c}{Cross-Reenactment} \\
        \cmidrule(lr){2-8}
        \cmidrule(lr){9-10}
        & SSIM$\uparrow$ & LPIPS$\downarrow$ & PSNR$\uparrow$ & CSIM$\uparrow$
        & AKD$\downarrow$ & FID$\downarrow$ & FVD$\downarrow$
        & CSIM$\uparrow$ & IQA$\uparrow$ \\
        \midrule
        w/ SHeaP
            & 0.704 & 0.188 & 20.67 & 0.860
            & 3.824 & 21.57 & \textbf{0.017}
            & 0.297 & 61.26 \\
        w/ Pixel3DMM
            & \textbf{0.722} & \textbf{0.178} & \textbf{21.19} & \textbf{0.879}
            & \textbf{2.752} & \textbf{21.29} & 0.020
            & \textbf{0.372} & \textbf{62.85} \\
        \bottomrule
    \end{tabular}
    }
    \caption{\textbf{Tracking-quality sensitivity on VFHQ using SHeaP or Pixel3DMM conditioning.}}
    \label{tab:sheap_test}
\end{table}

\paragraph{Discussion.}
The unified conditioning layer gives the best overall architectural trade-off with fewer parameters. Pixel-aligned normals provide the strongest overall representation, although alternatives lead on a few metrics. The CFG branches act as identity anchor, geometry driver, and semantic enhancer: image and geometry control identity and motion, while text improves semantic and perceptual balance.

%
%

\section{Conclusion}
\label{sec:conclu}
We presented ViDS, a Video Diffusion Shader for portrait animation conditioned on dense Pixel3DMM normal maps. Fixed reference identity parameters separate identity shape from driving pose and expression, while unified channel conditioning and multi-branch CFG control geometry, identity, and style. ViDS achieves the best overall performance among the evaluated baselines, leads on most self- and cross-reenactment metrics, generalizes to in-the-wild references, and aligns with user judgments.

Limitations include tracking sensitivity, slow autoregressive inference, no relighting, and limited control beyond 3DMM regions. Future work will pursue faster sampling and richer conditioning. We target technical analysis and creative tools, not deception.

\clearpage
\restorestandardlayout
{
    \small
    \bibliographystyle{ieeenat_fullname}
    \bibliography{main}

@String(CVPR  = {IEEE Conf. Comput. Vis. Pattern Recog.})

@String(ICCV  = {Int. Conf. Comput. Vis.})

@String(ECCV  = {Eur. Conf. Comput. Vis.})

@String(NeurIPS = {Adv. Neural Inform. Process. Syst.})

@String(CVPRW = {IEEE Conf. Comput. Vis. Pattern Recog. Worksh.})

@String(AAAI  = {AAAI})

@String(TOG   = {ACM Trans. Graph.})

@String(CVPR  = {CVPR})

@String(ICCV  = {ICCV})

@String(ECCV  = {ECCV})

@String(NeurIPS = {NeurIPS})

@String(CVPRW = {CVPRW})

@String(TOG   = {ACM TOG})

@inproceedings{booth20173d,
  title={3d face morphable models" in-the-wild"},
  author={Booth, James and Antonakos, Epameinondas and Ploumpis, Stylianos and Trigeorgis, George and Panagakis, Yannis and Zafeiriou, Stefanos},
  booktitle={Proceedings of the IEEE conference on computer vision and pattern recognition},
  pages={48--57},
  year={2017}
}

@inproceedings{booth20163d,
  title={A 3d morphable model learnt from 10,000 faces},
  author={Booth, James and Roussos, Anastasios and Zafeiriou, Stefanos and Ponniah, Allan and Dunaway, David},
  booktitle={Proceedings of the IEEE conference on computer vision and pattern recognition},
  pages={5543--5552},
  year={2016}
}

@inproceedings{ploumpis2019combining,
  title={Combining 3d morphable models: A large scale face-and-head model},
  author={Ploumpis, Stylianos and Wang, Haoyang and Pears, Nick and Smith, William AP and Zafeiriou, Stefanos},
  booktitle={Proceedings of the IEEE/CVF Conference on Computer Vision and Pattern Recognition},
  pages={10934--10943},
  year={2019}
}

@inproceedings{giebenhain2024mononphm,
 author={Simon Giebenhain and Tobias Kirschstein and Markos Georgopoulos and  Martin R{\"{u}}nz and Lourdes Agapito and Matthias Nie{\ss}ner},
 title={MonoNPHM: Dynamic Head Reconstruction from Monocular Videos},
 booktitle = {Proc. IEEE Conf. on Computer Vision and Pattern Recognition (CVPR)},
 year = {2024}}

@inproceedings{yenamandra2021i3dmm,
  title={i3DMM: Deep Implicit 3D Morphable Model of Human Heads},
  author={Tarun Yenamandra and Ayush Tewari and Florian Bernard and Hans-Peter Seidel and Mohamed Elgharib and Daniel Cremers and Christian Theobalt},
  booktitle={CVPR},
  year={2021}
}

@inproceedings{zheng2022imface,
  title={Imface: A nonlinear 3d morphable face model with implicit neural representations},
  author={Zheng, Mingwu and Yang, Hongyu and Huang, Di and Chen, Liming},
  booktitle={Proceedings of the IEEE/CVF Conference on Computer Vision and Pattern Recognition},
  pages={20343--20352},
  year={2022}
}

@inproceedings{bolkart2015groupwise,
  title={A groupwise multilinear correspondence optimization for 3d faces},
  author={Bolkart, Timo and Wuhrer, Stefanie},
  booktitle={Proceedings of the IEEE international conference on computer vision},
  pages={3604--3612},
  year={2015}
}

@inproceedings{brunton2014multilinear,
  title={Multilinear wavelets: A statistical shape space for human faces},
  author={Brunton, Alan and Bolkart, Timo and Wuhrer, Stefanie},
  booktitle={Computer Vision--ECCV 2014: 13th European Conference, Zurich, Switzerland, September 6-12, 2014, Proceedings, Part I 13},
  pages={297--312},
  year={2014},
  organization={Springer}
}

@inproceedings{tran2018nonlinear,
  title={Nonlinear 3d face morphable model},
  author={Tran, Luan and Liu, Xiaoming},
  booktitle={Proceedings of the IEEE conference on computer vision and pattern recognition},
  pages={7346--7355},
  year={2018}
}

@article{li2017learning,
  title={Learning a model of facial shape and expression from 4D scans.},
  author={Li, Tianye and Bolkart, Timo and Black, Michael J and Li, Hao and Romero, Javier},
  journal={ACM Trans. Graph.},
  volume={36},
  number={6},
  pages={194--1},
  year={2017}
}

@article{cao2013facewarehouse,
  title={Facewarehouse: A 3d facial expression database for visual computing},
  author={Cao, Chen and Weng, Yanlin and Zhou, Shun and Tong, Yiying and Zhou, Kun},
  journal={IEEE Transactions on Visualization and Computer Graphics},
  volume={20},
  number={3},
  pages={413--425},
  year={2013},
  publisher={IEEE}
}

@inproceedings{wood2022dense,
  title={3d face reconstruction with dense landmarks},
  author={Wood, Erroll and Baltru{\v{s}}aitis, Tadas and Hewitt, Charlie and Johnson, Matthew and Shen, Jingjing and Milosavljevi{\'c}, Nikola and Wilde, Daniel and Garbin, Stephan and Sharp, Toby and Stojiljkovi{\'c}, Ivan and others},
  booktitle={European Conference on Computer Vision},
  pages={160--177},
  year={2022},
  organization={Springer}
}

@InProceedings{taubner2024flowface,
  author    = {Taubner, Felix and Raina, Prashant and Tuli, Mathieu and Teh, Eu Wern and Lee, Chul and Huang, Jinmiao},
  title     = {{3D} Face Tracking from {2D} Video through Iterative Dense {UV} to Image Flow},
  booktitle = {Proceedings of the IEEE/CVF Conference on Computer Vision and Pattern Recognition (CVPR)},
  month     = {June},
  year      = {2024},
  pages     = {1227-1237}
}

@misc{giebenhain2025pixel3dmm,
title={Pixel3DMM: Versatile Screen-Space Priors for Single-Image 3D Face Reconstruction},
author={Simon Giebenhain and Tobias Kirschstein and  Martin R{\"{u}}nz and Lourdes Agapito and Matthias Nie{\ss}ner},
year={2025},
url={https://arxiv.org/abs/2505.00615},
}

@article{Thies2016Face2FaceRF,
  title={Face2Face: Real-Time Face Capture and Reenactment of RGB Videos},
  author={Justus Thies and Michael Zollh{\"o}fer and Marc Stamminger and Christian Theobalt and Matthias Nie{\ss}ner},
  journal={2016 IEEE Conference on Computer Vision and Pattern Recognition (CVPR)},
  year={2016},
  pages={2387-2395},
  url={https://api.semanticscholar.org/CorpusID:52858569}
}

@article{Deng2019Accurate3F,
  title={Accurate 3D Face Reconstruction With Weakly-Supervised Learning: From Single Image to Image Set},
  author={Yu Deng and Jiaolong Yang and Sicheng Xu and Dong Chen and Yunde Jia and Xin Tong},
  journal={2019 IEEE/CVF Conference on Computer Vision and Pattern Recognition Workshops (CVPRW)},
  year={2019},
  pages={285-295},
  url={https://api.semanticscholar.org/CorpusID:84187285}
}

@article{Sanyal2019LearningTR,
  title={Learning to Regress 3D Face Shape and Expression From an Image Without 3D Supervision},
  author={Soubhik Sanyal and Timo Bolkart and Haiwen Feng and Michael J. Black},
  journal={2019 IEEE/CVF Conference on Computer Vision and Pattern Recognition (CVPR)},
  year={2019},
  pages={7755-7764},
  url={https://api.semanticscholar.org/CorpusID:155100135}
}

@article{Feng2020LearningAA,
  title={Learning an animatable detailed 3D face model from in-the-wild images},
  author={Yao Feng and Haiwen Feng and Michael J. Black and Timo Bolkart},
  journal={ACM Transactions on Graphics (TOG)},
  year={2020},
  volume={40},
  pages={1 - 13},
  url={https://api.semanticscholar.org/CorpusID:236094976}
}

@inproceedings{Danek2022EMOCAED,
  title={Emoca: Emotion driven monocular face capture and animation},
  author={Dan{\v{e}}{\v{c}}ek, Radek and Black, Michael J and Bolkart, Timo},
  booktitle={Proceedings of the IEEE/CVF Conference on Computer Vision and Pattern Recognition},
  pages={20311--20322},
  year={2022}
}

@inproceedings{Zielonka2022TowardsMR,
  title={Towards Metrical Reconstruction of Human Faces},
  author={Wojciech Zielonka and Timo Bolkart and Justus Thies},
  booktitle={European Conference on Computer Vision},
  year={2022},
  url={https://api.semanticscholar.org/CorpusID:248177832}
}

@inproceedings{Retsinas20243DFE,
  title={3d facial expressions through analysis-by-neural-synthesis},
  author={Retsinas, George and Filntisis, Panagiotis P and Danecek, Radek and Abrevaya, Victoria F and Roussos, Anastasios and Bolkart, Timo and Maragos, Petros},
  booktitle={Proceedings of the IEEE/CVF Conference on Computer Vision and Pattern Recognition},
  pages={2490--2501},
  year={2024}
}

@article{Schoneveld2025SHeaPSH,
  title={SHeaP: Self-Supervised Head Geometry Predictor Learned via 2D Gaussians},
  author={Liam Schoneveld and Zhe Chen and Davide Davoli and Jiapeng Tang and Saimon Terazawa and Ko Nishino and Matthias Nie{\ss}ner},
  journal={ArXiv},
  year={2025},
  volume={abs/2504.12292},
  url={https://api.semanticscholar.org/CorpusID:277823993}
}

@article{he2022latent,
  title={Latent video diffusion models for high-fidelity long video generation},
  author={He, Yingqing and Yang, Tianyu and Zhang, Yong and Shan, Ying and Chen, Qifeng},
  journal={arXiv preprint arXiv:2211.13221},
  year={2022}
}

@article{text2video-zero,
    title={Text2Video-Zero: Text-to-Image Diffusion Models are Zero-Shot Video Generators},
    author={Khachatryan, Levon and Movsisyan, Andranik and Tadevosyan, Vahram and Henschel, Roberto and Wang, Zhangyang and Navasardyan, Shant and Shi, Humphrey},
    journal={arXiv preprint arXiv:2303.13439},
    year={2023}
  }

@article{xing2023make,
					  author  = {Xing, Jinbo and Xia, Menghan and Liu, Yuxin and Zhang, Yuechen and Zhang, Yong and He, Yingqing and Liu, Hanyuan and Chen, Haoxin and Cun, Xiaodong and Wang, Xintao and Shan, Ying and Wong, Tien-Tsin},
					  title   = {Make-Your-Video: Customized Video Generation Using Textual and Structural Guidance},
					  journal = {arXiv preprint arXiv:2306.00943},
					  year    = {2023}
					}

@misc{chen2023videodreamer,
          title={VideoDreamer: Customized Multi-Subject Text-to-Video Generation with Disen-Mix Finetuning},
          author={Hong Chen and Xin Wang and Guanning Zeng and Yipeng Zhang and Yuwei Zhou and Feilin Han and Wenwu Zhu},
          year={2023},
          eprint={2311.00990},
          archivePrefix={arXiv},
          primaryClass={cs.CV}
    }

@inproceedings{rombach2022highstable_diffusion,
  title={High-resolution image synthesis with latent diffusion models},
  author={Rombach, Robin and Blattmann, Andreas and Lorenz, Dominik and Esser, Patrick and Ommer, Bj{\"o}rn},
  booktitle={Proceedings of the IEEE/CVF conference on computer vision and pattern recognition},
  pages={10684--10695},
  year={2022}
}

@article{mildenhall2021nerf,
  title={Nerf: Representing scenes as neural radiance fields for view synthesis},
  author={Mildenhall, Ben and Srinivasan, Pratul P and Tancik, Matthew and Barron, Jonathan T and Ramamoorthi, Ravi and Ng, Ren},
  journal={Communications of the ACM},
  volume={65},
  number={1},
  pages={99--106},
  year={2021},
  publisher={ACM New York, NY, USA}
}

@inproceedings{giebenhain2023learning,
  title={Learning Neural Parametric Head Models},
  author={Giebenhain, Simon and Kirschstein, Tobias and Georgopoulos, Markos and R{\"u}nz, Martin and Agapito, Lourdes and Nie{\ss}ner, Matthias},
  booktitle={Proceedings of the IEEE/CVF Conference on Computer Vision and Pattern Recognition},
  pages={21003--21012},
  year={2023}
}

@article{Ho2022VideoDM,
  title={Video Diffusion Models},
  author={Jonathan Ho and Tim Salimans and Alexey Gritsenko and William Chan and Mohammad Norouzi and David J. Fleet},
  journal={ArXiv},
  year={2022},
  volume={abs/2204.03458},
  url={https://api.semanticscholar.org/CorpusID:248006185}
}

@article{Tewari2020StateOT,
  title={State of the Art on Neural Rendering},
  author={Ayush Kumar Tewari and Ohad Fried and Justus Thies and Vincent Sitzmann and Stephen Lombardi and Kalyan Sunkavalli and Ricardo Martin-Brualla and Tomas Simon and Jason M. Saragih and Matthias Nie{\ss}ner and Rohit Pandey and S. Fanello and Gordon Wetzstein and Jun-Yan Zhu and Christian Theobalt and Maneesh Agrawala and Eli Shechtman and Dan B. Goldman and Michael Zollhofer},
  journal={Computer Graphics Forum},
  year={2020},
  volume={39},
  url={https://api.semanticscholar.org/CorpusID:215416317}
}

@inproceedings{park2019deepsdf,
  title={Deepsdf: Learning continuous signed distance functions for shape representation},
  author={Park, Jeong Joon and Florence, Peter and Straub, Julian and Newcombe, Richard and Lovegrove, Steven},
  booktitle={Proceedings of the IEEE/CVF conference on computer vision and pattern recognition},
  pages={165--174},
  year={2019}
}

@ARTICLE{zhe2022ibrdf,
author={Chen, Zhe and Nobuhara, Shohei and Nishino, Ko},
journal={ IEEE Transactions on Pattern Analysis \& Machine Intelligence },
title={{ Invertible Neural BRDF for Object Inverse Rendering }},
year={2022},
volume={44},
number={12},
ISSN={1939-3539},
pages={9380-9395},
doi={10.1109/TPAMI.2021.3129537},
url = {https://doi.ieeecomputersociety.org/10.1109/TPAMI.2021.3129537},
publisher={IEEE Computer Society},
address={Los Alamitos, CA, USA},
month=dec}

@article{kerbl20233d,
  title={3D Gaussian splatting for real-time radiance field rendering.},
  author={Kerbl, Bernhard and Kopanas, Georgios and Leimk{\"u}hler, Thomas and Drettakis, George},
  journal={ACM Trans. Graph.},
  volume={42},
  number={4},
  pages={139--1},
  year={2023}
}

@article{muller2022instant,
  title={Instant neural graphics primitives with a multiresolution hash encoding},
  author={M{\"u}ller, Thomas and Evans, Alex and Schied, Christoph and Keller, Alexander},
  journal={ACM transactions on graphics (TOG)},
  volume={41},
  number={4},
  pages={1--15},
  year={2022},
  publisher={ACM New York, NY, USA}
}

@inproceedings {zeng2025renderformer,
    title      = {RenderFormer: Transformer-based Neural Rendering of Triangle Meshes with Global Illumination},
    author     = {Chong Zeng and Yue Dong and Pieter Peers and Hongzhi Wu and Xin Tong},
    booktitle  = {ACM SIGGRAPH 2025 Conference Papers},
    year       = {2025}
}

@article{gao2020deferred,
author = {Gao, Duan and Chen, Guojun and Dong, Yue and Peers, Pieter and Xu, Kun and Tong, Xin},
title = {Deferred neural lighting: free-viewpoint relighting from unstructured photographs},
year = {2020},
issue_date = {December 2020},
publisher = {Association for Computing Machinery},
address = {New York, NY, USA},
volume = {39},
number = {6},
issn = {0730-0301},
url = {https://doi.org/10.1145/3414685.3417767},
doi = {10.1145/3414685.3417767},
journal = {ACM Trans. Graph.},
month = nov,
articleno = {258},
numpages = {15}
}

@article{thies2019deferred,
author = {Thies, Justus and Zollh\"{o}fer, Michael and Nie\ss{}ner, Matthias},
title = {Deferred neural rendering: image synthesis using neural textures},
year = {2019},
issue_date = {August 2019},
publisher = {Association for Computing Machinery},
address = {New York, NY, USA},
volume = {38},
number = {4},
issn = {0730-0301},
url = {https://doi.org/10.1145/3306346.3323035},
doi = {10.1145/3306346.3323035},
journal = {ACM Trans. Graph.},
month = jul,
articleno = {66},
numpages = {12}
}

@inproceedings{aliev2020neural,
  title={Neural point-based graphics},
  author={Aliev, Kara-Ali and Sevastopolsky, Artem and Kolos, Maria and Ulyanov, Dmitry and Lempitsky, Victor},
  booktitle={European conference on computer vision},
  pages={696--712},
  year={2020},
  organization={Springer}
}

@inproceedings{DiffusionRenderer,
    author = {Ruofan Liang and Zan Gojcic and Huan Ling and Jacob Munkberg and
        Jon Hasselgren and Zhi-Hao Lin and Jun Gao and Alexander Keller and
        Nandita Vijaykumar and Sanja Fidler and Zian Wang},
    title = {DiffusionRenderer: Neural Inverse and Forward Rendering with Video Diffusion Models},
    booktitle = {The IEEE Conference on Computer Vision and Pattern Recognition (CVPR)},
    month = {June},
    year = {2025}
}

@inproceedings{ma2024followyourpose,
  title={Follow your pose: Pose-guided text-to-video generation using pose-free videos},
  author={Ma, Yue and He, Yingqing and Cun, Xiaodong and Wang, Xintao and Chen, Siran and Li, Xiu and Chen, Qifeng},
  booktitle={Proceedings of the AAAI Conference on Artificial Intelligence},
  volume={38},
  number={5},
  pages={4117--4125},
  year={2024}
}

@article{Gan2025HumanDiTPD,
  title={HumanDiT: Pose-Guided Diffusion Transformer for Long-form Human Motion Video Generation},
  author={Qijun Gan and Yi Ren and Chen Zhang and Zhenhui Ye and Pan Xie and Xiang Yin and Zehuan Yuan and Bingyue Peng and Jianke Zhu},
  journal={ArXiv},
  year={2025},
  volume={abs/2502.04847},
  url={https://api.semanticscholar.org/CorpusID:276236107}
}

@article{Guo2024LivePortraitEP,
  title={LivePortrait: Efficient Portrait Animation with Stitching and Retargeting Control},
  author={Jianzhu Guo and Dingyun Zhang and Xiaoqiang Liu and Zhizhou Zhong and Yuan Zhang and Pengfei Wan and Dingyun Zhang},
  journal={ArXiv},
  year={2024},
  volume={abs/2407.03168},
  url={https://api.semanticscholar.org/CorpusID:270924349}
}

@article{Hao2021GANcraftU3,
  title={GANcraft: Unsupervised 3D Neural Rendering of Minecraft Worlds},
  author={Zekun Hao and Arun Mallya and Serge J. Belongie and Ming-Yu Liu},
  journal={2021 IEEE/CVF International Conference on Computer Vision (ICCV)},
  year={2021},
  pages={14052-14062},
  url={https://api.semanticscholar.org/CorpusID:233240846}
}

@inproceedings{SadTalker,
  author       = {Wenxuan Zhang and
                  Xiaodong Cun and
                  Xuan Wang and
                  Yong Zhang and
                  Xi Shen and
                  Yu Guo and
                  Ying Shan and
                  Fei Wang},
  title        = {SadTalker: Learning Realistic 3D Motion Coefficients for Stylized
                  Audio-Driven Single Image Talking Face Animation},
  booktitle    = {{IEEE/CVF} Conference on Computer Vision and Pattern Recognition,
                  {CVPR} 2023, Vancouver, BC, Canada, June 17-24, 2023},
  pages        = {8652--8661},
  publisher    = {{IEEE}},
  year         = {2023},
  url          = {https://doi.org/10.1109/CVPR52729.2023.00836},
  doi          = {10.1109/CVPR52729.2023.00836},
  bibsource    = {dblp computer science bibliography, https://dblp.org}
}

@article{abs-2211-12368,
  author       = {Jiaxiang Tang and
                  Kaisiyuan Wang and
                  Hang Zhou and
                  Xiaokang Chen and
                  Dongliang He and
                  Tianshu Hu and
                  Jingtuo Liu and
                  Gang Zeng and
                  Jingdong Wang},
  title        = {Real-time Neural Radiance Talking Portrait Synthesis via Audio-spatial
                  Decomposition},
  journal      = {CoRR},
  volume       = {abs/2211.12368},
  year         = {2022},
  url          = {https://doi.org/10.48550/arXiv.2211.12368},
  doi          = {10.48550/ARXIV.2211.12368},
  eprinttype    = {arXiv},
  eprint       = {2211.12368},
  bibsource    = {dblp computer science bibliography, https://dblp.org}
}

@inproceedings{GuoCLLBZ21,
  author       = {Yudong Guo and
                  Keyu Chen and
                  Sen Liang and
                  Yong{-}Jin Liu and
                  Hujun Bao and
                  Juyong Zhang},
  title        = {AD-NeRF: Audio Driven Neural Radiance Fields for Talking Head Synthesis},
  booktitle    = {2021 {IEEE/CVF} International Conference on Computer Vision, {ICCV}
                  2021, Montreal, QC, Canada, October 10-17, 2021},
  pages        = {5764--5774},
  publisher    = {{IEEE}},
  year         = {2021},
  url          = {https://doi.org/10.1109/ICCV48922.2021.00573},
  doi          = {10.1109/ICCV48922.2021.00573},
  bibsource    = {dblp computer science bibliography, https://dblp.org}
}

@inproceedings{BlanzV99,
  author       = {Volker Blanz and
                  Thomas Vetter},
  editor       = {Warren N. Waggenspack},
  title        = {A Morphable Model for the Synthesis of 3D Faces},
  booktitle    = {Proceedings of the 26th Annual Conference on Computer Graphics and
                  Interactive Techniques, {SIGGRAPH} 1999, Los Angeles, CA, USA, August
                  8-13, 1999},
  pages        = {187--194},
  publisher    = {{ACM}},
  year         = {1999},
  url          = {https://dl.acm.org/citation.cfm?id=311556},
  bibsource    = {dblp computer science bibliography, https://dblp.org}
}

@article{wan2025,
      title={Wan: Open and Advanced Large-Scale Video Generative Models},
      author={Team Wan and Ang Wang and Baole Ai and Bin Wen and Chaojie Mao and Chen-Wei Xie and Di Chen and Feiwu Yu and Haiming Zhao and Jianxiao Yang and Jianyuan Zeng and Jiayu Wang and Jingfeng Zhang and Jingren Zhou and Jinkai Wang and Jixuan Chen and Kai Zhu and Kang Zhao and Keyu Yan and Lianghua Huang and Mengyang Feng and Ningyi Zhang and Pandeng Li and Pingyu Wu and Ruihang Chu and Ruili Feng and Shiwei Zhang and Siyang Sun and Tao Fang and Tianxing Wang and Tianyi Gui and Tingyu Weng and Tong Shen and Wei Lin and Wei Wang and Wei Wang and Wenmeng Zhou and Wente Wang and Wenting Shen and Wenyuan Yu and Xianzhong Shi and Xiaoming Huang and Xin Xu and Yan Kou and Yangyu Lv and Yifei Li and Yijing Liu and Yiming Wang and Yingya Zhang and Yitong Huang and Yong Li and You Wu and Yu Liu and Yulin Pan and Yun Zheng and Yuntao Hong and Yupeng Shi and Yutong Feng and Zeyinzi Jiang and Zhen Han and Zhi-Fan Wu and Ziyu Liu},
      journal = {arXiv preprint arXiv:2503.20314},
      year={2025}
}

@inproceedings{DiT,
  title={Scalable diffusion models with transformers},
  author={Peebles, William and Xie, Saining},
  booktitle={Proceedings of the IEEE/CVF international conference on computer vision},
  pages={4195--4205},
  year={2023}
}

@article{flow_matching,
  title={Flow matching for generative modeling},
  author={Lipman, Yaron and Chen, Ricky TQ and Ben-Hamu, Heli and Nickel, Maximilian and Le, Matt},
  journal={arXiv preprint arXiv:2210.02747},
  year={2022}
}

@article{xu2024hallo,
  title={Hallo: Hierarchical audio-driven visual synthesis for portrait image animation},
  author={Xu, Mingwang and Li, Hui and Su, Qingkun and Shang, Hanlin and Zhang, Liwei and Liu, Ce and Wang, Jingdong and Yao, Yao and Zhu, Siyu},
  journal={arXiv preprint arXiv:2406.08801},
  year={2024}
}

@article{chu2024Gagavatar,
  title={Generalizable and animatable gaussian head avatar},
  author={Chu, Xuangeng and Harada, Tatsuya},
  journal={Advances in Neural Information Processing Systems},
  volume={37},
  pages={57642--57670},
  year={2024}
}

@inproceedings{xie2022vfhq,
  title={Vfhq: A high-quality dataset and benchmark for video face super-resolution},
  author={Xie, Liangbin and Wang, Xintao and Zhang, Honglun and Dong, Chao and Shan, Ying},
  booktitle={Proceedings of the IEEE/CVF Conference on Computer Vision and Pattern Recognition},
  pages={657--666},
  year={2022}
}

@inproceedings{zhang2021hdtf,
  title={Flow-guided one-shot talking face generation with a high-resolution audio-visual dataset},
  author={Zhang, Zhimeng and Li, Lincheng and Ding, Yu and Fan, Changjie},
  booktitle={Proceedings of the IEEE/CVF conference on computer vision and pattern recognition},
  pages={3661--3670},
  year={2021}
}

@article{kirschstein2023nersemble,
    author = {Kirschstein, Tobias and Qian, Shenhan and Giebenhain, Simon and Walter, Tim and Nie\ss{}ner, Matthias},
    title = {NeRSemble: Multi-View Radiance Field Reconstruction of Human Heads},
    year = {2023},
    issue_date = {August 2023},
    publisher = {Association for Computing Machinery},
    address = {New York, NY, USA},
    volume = {42},
    number = {4},
    issn = {0730-0301},
    url = {https://doi.org/10.1145/3592455},
    doi = {10.1145/3592455},
    journal = {ACM Trans. Graph.},
    month = {jul},
    articleno = {161},
    numpages = {14},
}

@inproceedings{zhang2018lpips,
  title={The unreasonable effectiveness of deep features as a perceptual metric},
  author={Zhang, Richard and Isola, Phillip and Efros, Alexei A and Shechtman, Eli and Wang, Oliver},
  booktitle={Proceedings of the IEEE conference on computer vision and pattern recognition},
  pages={586--595},
  year={2018}
}

@article{wang2004SSIM,
  title={Image quality assessment: from error visibility to structural similarity},
  author={Wang, Zhou and Bovik, Alan C and Sheikh, Hamid R and Simoncelli, Eero P},
  journal={IEEE transactions on image processing},
  volume={13},
  number={4},
  pages={600--612},
  year={2004},
  publisher={IEEE}
}

@inproceedings{deng2019arcface,
  title={Arcface: Additive angular margin loss for deep face recognition},
  author={Deng, Jiankang and Guo, Jia and Xue, Niannan and Zafeiriou, Stefanos},
  booktitle={Proceedings of the IEEE/CVF conference on computer vision and pattern recognition},
  pages={4690--4699},
  year={2019}
}

@inproceedings{deng2019accurate,
  title={Accurate 3d face reconstruction with weakly-supervised learning: From single image to image set},
  author={Deng, Yu and Yang, Jiaolong and Xu, Sicheng and Chen, Dong and Jia, Yunde and Tong, Xin},
  booktitle={Proceedings of the IEEE/CVF conference on computer vision and pattern recognition workshops},
  pages={0--0},
  year={2019}
}

@inproceedings{bulat2017far,
  title={How far are we from solving the 2d \& 3d face alignment problem?(and a dataset of 230,000 3d facial landmarks)},
  author={Bulat, Adrian and Tzimiropoulos, Georgios},
  booktitle={Proceedings of the IEEE international conference on computer vision},
  pages={1021--1030},
  year={2017}
}

@inproceedings{ma2024follow-your-emoji,
  title={Follow-your-emoji: Fine-controllable and expressive freestyle portrait animation},
  author={Ma, Yue and Liu, Hongyu and Wang, Hongfa and Pan, Heng and He, Yingqing and Yuan, Junkun and Zeng, Ailing and Cai, Chengfei and Shum, Heung-Yeung and Liu, Wei and others},
  booktitle={SIGGRAPH Asia 2024 Conference Papers},
  pages={1--12},
  year={2024}
}

@inproceedings{xie2024xportrait,
  title={X-portrait: Expressive portrait animation with hierarchical motion attention},
  author={Xie, You and Xu, Hongyi and Song, Guoxian and Wang, Chao and Shi, Yichun and Luo, Linjie},
  booktitle={ACM SIGGRAPH 2024 Conference Papers},
  pages={1--11},
  year={2024}
}

@inproceedings{xu2025hunyuanportrait,
  title={Hunyuanportrait: Implicit condition control for enhanced portrait animation},
  author={Xu, Zunnan and Yu, Zhentao and Zhou, Zixiang and Zhou, Jun and Jin, Xiaoyu and Hong, Fa-Ting and Ji, Xiaozhong and Zhu, Junwei and Cai, Chengfei and Tang, Shiyu and others},
  booktitle={Proceedings of the Computer Vision and Pattern Recognition Conference},
  pages={15909--15919},
  year={2025}
}

@inproceedings{he2025lam,
  title={LAM: Large Avatar Model for One-shot Animatable Gaussian Head},
  author={He, Yisheng and Gu, Xiaodong and Ye, Xiaodan and Xu, Chao and Zhao, Zhengyi and Dong, Yuan and Yuan, Weihao and Dong, Zilong and Bo, Liefeng},
  booktitle={Proceedings of the Special Interest Group on Computer Graphics and Interactive Techniques Conference Conference Papers},
  pages={1--13},
  year={2025}
}

@InProceedings{Siarohin_2019_NeurIPS,
  author={Siarohin, Aliaksandr and Lathuilière, Stéphane and Tulyakov, Sergey and Ricci, Elisa and Sebe, Nicu},
  title={First Order Motion Model for Image Animation},
  booktitle = {Conference on Neural Information Processing Systems (NeurIPS)},
  month = {December},
  year = {2019}
}

@inproceedings{wang2021facevid2vid,
  title={One-Shot Free-View Neural Talking-Head Synthesis for Video Conferencing},
  author={Ting-Chun Wang and Arun Mallya and Ming-Yu Liu},
  booktitle={Proceedings of the IEEE Conference on Computer Vision and Pattern Recognition},
  year={2021}
}

@inproceedings{Drobyshev22MP,
  title={Megaportraits: One-shot megapixel neural head avatars},
  author={Drobyshev, Nikita and Chelishev, Jenya and Khakhulin, Taras and Ivakhnenko, Aleksei and Lempitsky, Victor and Zakharov, Egor},
  booktitle={Proceedings of the 30th ACM International Conference on Multimedia},
  pages={2663--2671},
  year={2022}
}

@article{Xu2024VASA1LA,
  title={VASA-1: Lifelike Audio-Driven Talking Faces Generated in Real Time},
  author={Sicheng Xu and Guojun Chen and Yufeng Guo and Jiaolong Yang and Chong Li and Zhenyu Zang and Yizhong Zhang and Xin Tong and Baining Guo},
  journal={ArXiv},
  year={2024},
  volume={abs/2404.10667},
  url={https://api.semanticscholar.org/CorpusID:269157309}
}

@inproceedings{Zhou2021Pose,
  title = {Pose-Controllable Talking Face Generation by Implicitly Modularized Audio-Visual Representation},
  author = {Zhou, Hang and Sun, Yasheng and Wu, Wayne and Loy, Chen Change and Wang, Xiaogang and Liu, Ziwei},
  booktitle = {Proceedings of the IEEE Conference on Computer Vision and Pattern Recognition (CVPR)},
  year = {2021}
}

@article{Zhang2021FlowguidedOT,
  title={Flow-guided One-shot Talking Face Generation with a High-resolution Audio-visual Dataset},
  author={Zhimeng Zhang and Lincheng Li and Yu Ding},
  journal={2021 IEEE/CVF Conference on Computer Vision and Pattern Recognition (CVPR)},
  year={2021},
  pages={3660-3669},
  url={https://api.semanticscholar.org/CorpusID:235657610}
}

@inproceedings{zhang2024personatalk,
      title={Personatalk: Bring attention to your persona in visual dubbing},
      author={Zhang, Longhao and Liang, Shuang and Ge, Zhipeng and Hu, Tianshu},
      booktitle={SIGGRAPH Asia 2024 Conference Papers},
      pages={1--9},
      year={2024}
}

@article{thies2020nvp,
  author = {Thies, Justus and Elgharib, Mohamed and Tewari, Ayush and Theobalt, Christian and Nie{\ss}ner, Matthias},
  title = {Neural Voice Puppetry: Audio-driven Facial Reenactment},
  journal={ECCV 2020},
  year={2020}
}

@inproceedings{INSTA:CVPR2023,
  title={Instant volumetric head avatars},
  author={Zielonka, Wojciech and Bolkart, Timo and Thies, Justus},
  booktitle={Proceedings of the IEEE/CVF conference on computer vision and pattern recognition},
  pages={4574--4584},
  year={2023}
}

@inproceedings{gafni2021dynamic,
  title={Dynamic neural radiance fields for monocular 4d facial avatar reconstruction},
  author={Gafni, Guy and Thies, Justus and Zollhofer, Michael and Nie{\ss}ner, Matthias},
  booktitle={Proceedings of the IEEE/CVF Conference on Computer Vision and Pattern Recognition},
  pages={8649--8658},
  year={2021}
}

@article{gao2022reconstructing,
  title={Reconstructing personalized semantic facial nerf models from monocular video},
  author={Gao, Xuan and Zhong, Chenglai and Xiang, Jun and Hong, Yang and Guo, Yudong and Zhang, Juyong},
  journal={ACM Transactions on Graphics (TOG)},
  volume={41},
  number={6},
  pages={1--12},
  year={2022},
  publisher={ACM New York, NY, USA}
}

@article{Qian2023GaussianAvatarsPH,
  title={GaussianAvatars: Photorealistic Head Avatars with Rigged 3D Gaussians},
  author={Shenhan Qian and Tobias Kirschstein and Liam Schoneveld and Davide Davoli and Simon Giebenhain and Matthias Nie{\ss}ner},
  journal={2024 IEEE/CVF Conference on Computer Vision and Pattern Recognition (CVPR)},
  year={2023},
  pages={20299-20309},
  url={https://api.semanticscholar.org/CorpusID:265609441}
}

@article{Xiang2023FlashAvatarHH,
  title={FlashAvatar: High-Fidelity Head Avatar with Efficient Gaussian Embedding},
  author={Jun Xiang and Xuan Gao and Yudong Guo and Juyong Zhang},
  journal={2024 IEEE/CVF Conference on Computer Vision and Pattern Recognition (CVPR)},
  year={2023},
  pages={1802-1812},
  url={https://api.semanticscholar.org/CorpusID:268793428}
}

@inproceedings{xu2024gaussian,
  title={Gaussian head avatar: Ultra high-fidelity head avatar via dynamic gaussians},
  author={Xu, Yuelang and Chen, Benwang and Li, Zhe and Zhang, Hongwen and Wang, Lizhen and Zheng, Zerong and Liu, Yebin},
  booktitle={Proceedings of the IEEE/CVF conference on computer vision and pattern recognition},
  pages={1931--1941},
  year={2024}
}

@inproceedings{hu2024gaussianavatar,
  title={Gaussianavatar: Towards realistic human avatar modeling from a single video via animatable 3d gaussians},
  author={Hu, Liangxiao and Zhang, Hongwen and Zhang, Yuxiang and Zhou, Boyao and Liu, Boning and Zhang, Shengping and Nie, Liqiang},
  booktitle={Proceedings of the IEEE/CVF conference on computer vision and pattern recognition},
  pages={634--644},
  year={2024}
}

@article{ho2022classifierfree,
  title={Classifier-free diffusion guidance},
  author={Ho, Jonathan and Salimans, Tim},
  journal={arXiv preprint arXiv:2207.12598},
  year={2022}
}

@article{lugaresi2019mediapipe,
  title={Mediapipe: A framework for building perception pipelines},
  author={Lugaresi, Camillo and Tang, Jiuqiang and Nash, Hadon and McClanahan, Chris and Uboweja, Esha and Hays, Michael and Zhang, Fan and Chang, Chuo-Ling and Yong, Ming Guang and Lee, Juhyun and others},
  journal={arXiv preprint arXiv:1906.08172},
  year={2019}
}

@inproceedings{tang2024dphms,
  title={Dphms: Diffusion parametric head models for depth-based tracking},
  author={Tang, Jiapeng and Dai, Angela and Nie, Yinyu and Markhasin, Lev and Thies, Justus and Nie{\ss}ner, Matthias},
  booktitle={Proceedings of the IEEE/CVF conference on computer vision and pattern recognition},
  pages={1111--1122},
  year={2024}
}

@article{chung2023unimax,
  title={Unimax: Fairer and more effective language sampling for large-scale multilingual pretraining},
  author={Chung, Hyung Won and Constant, Noah and Garcia, Xavier and Roberts, Adam and Tay, Yi and Narang, Sharan and Firat, Orhan},
  journal={arXiv preprint arXiv:2304.09151},
  year={2023}
}

@article{FLAME:SiggraphAsia2017,
  title = {Learning a model of facial shape and expression from {4D} scans},
  author = {Li, Tianye and Bolkart, Timo and Black, Michael. J. and Li, Hao and Romero, Javier},
  journal = {ACM Transactions on Graphics, (Proc. SIGGRAPH Asia)},
  volume = {36},
  number = {6},
  year = {2017},
  pages = {194:1--194:17},
  url = {https://doi.org/10.1145/3130800.3130813}
}

@article{guo2023animatediff,
  title={Animatediff: Animate your personalized text-to-image diffusion models without specific tuning},
  author={Guo, Yuwei and Yang, Ceyuan and Rao, Anyi and Liang, Zhengyang and Wang, Yaohui and Qiao, Yu and Agrawala, Maneesh and Lin, Dahua and Dai, Bo},
  journal={arXiv preprint arXiv:2307.04725},
  year={2023}
}

@PROCEEDINGS{bfm09,
            title={A 3D Face Model for Pose and Illumination Invariant Face Recognition},
            author={P. Paysan and R. Knothe and B. Amberg
                    and S. Romdhani and T. Vetter},
            journal={Proceedings of the 6th IEEE International Conference on Advanced Video and Signal based Surveillance (AVSS)
                 for Security, Safety and Monitoring in Smart Environments},
            organization={IEEE},
            year={2009},
            address     = {Genova, Italy},
            }

@inproceedings{thies2016face2face,
  title={Face2face: Real-time face capture and reenactment of rgb videos},
  author={Thies, Justus and Zollhofer, Michael and Stamminger, Marc and Theobalt, Christian and Nie{\ss}ner, Matthias},
  booktitle={Proceedings of the IEEE conference on computer vision and pattern recognition},
  pages={2387--2395},
  year={2016}
}

@article{thies2015real,
  title={Real-time expression transfer for facial reenactment.},
  author={Thies, Justus and Zollh{\"o}fer, Michael and Nie{\ss}ner, Matthias and Valgaerts, Levi and Stamminger, Marc and Theobalt, Christian},
  journal={ACM Trans. Graph.},
  volume={34},
  number={6},
  pages={183--1},
  year={2015}
}

@article{thies2018headon,
  title={Headon: Real-time reenactment of human portrait videos},
  author={Thies, Justus and Zollh{\"o}fer, Michael and Theobalt, Christian and Stamminger, Marc and Nie{\ss}ner, Matthias},
  journal={ACM Transactions on Graphics (TOG)},
  volume={37},
  number={4},
  pages={1--13},
  year={2018},
  publisher={ACM New York, NY, USA}
}

@article{wei2024aniportrait,
  title={Aniportrait: Audio-driven synthesis of photorealistic portrait animation},
  author={Wei, Huawei and Yang, Zejun and Wang, Zhisheng},
  journal={arXiv preprint arXiv:2403.17694},
  year={2024}
}

@inproceedings{chen2025echomimic,
  title={Echomimic: Lifelike audio-driven portrait animations through editable landmark conditions},
  author={Chen, Zhiyuan and Cao, Jiajiong and Chen, Zhiquan and Li, Yuming and Ma, Chenguang},
  booktitle={Proceedings of the AAAI Conference on Artificial Intelligence},
  volume={39},
  number={3},
  pages={2403--2410},
  year={2025}
}

@misc{prinzler2024joker,
      title={Joker: Conditional 3D Head Synthesis with Extreme Facial Expressions},
      author={Malte Prinzler and Egor Zakharov and Vanessa Sklyarova and Berna Kabadayi and Justus Thies},
      year={2024},
      eprint={2410.16395},
      archivePrefix={arXiv},
      primaryClass={cs.CV},
      url={https://arxiv.org/abs/2410.16395},
}

@inproceedings{tang2025gaf,
  title={Gaf: Gaussian avatar reconstruction from monocular videos via multi-view diffusion},
  author={Tang, Jiapeng and Davoli, Davide and Kirschstein, Tobias and Schoneveld, Liam and Niessner, Matthias},
  booktitle={Proceedings of the Computer Vision and Pattern Recognition Conference},
  pages={5546--5558},
  year={2025}
}

@article{cheng2025wananimate,
  title={Wan-animate: Unified character animation and replacement with holistic replication},
  author={Cheng, Gang and Gao, Xin and Hu, Li and Hu, Siqi and Huang, Mingyang and Ji, Chaonan and Li, Ju and Meng, Dechao and Qi, Jinwei and Qiao, Penchong and others},
  journal={arXiv preprint arXiv:2509.14055},
  year={2025}
}
}

\end{document}